\documentclass[11pt,a4paper]{article}
\usepackage[hyperref]{ranlp2023}
\usepackage{times}
\usepackage{latexsym}
\usepackage{csquotes}
\usepackage{booktabs}
\usepackage{makecell}
\usepackage{listings}
\usepackage[normalem]{ulem} 
\usepackage{tcolorbox}
\usepackage{siunitx}
\usepackage{multirow}
\usepackage{graphicx}
\definecolor{own_green}{HTML}{006400}
\definecolor{own_red}{HTML}{e61919}
\definecolor{own_grey}{HTML}{6e6e6e}
\definecolor{systembg}{rgb}{0.9, 0.9, 1}  
\definecolor{userbg}{rgb}{0.95, 1, 0.95}

\usepackage{microtype}

\aclfinalcopy 
\def\aclpaperid{128} 

\title{Simplifications are Absolutists: How Simplified Language Reduces Word Sense Awareness in LLM-Generated Definitions}

\author{Lukas Ellinger, Miriam Anschütz, and Georg Groh \\
  School for Computation, Information and Technology \\
  Technical University of Munich, Germany \\
  \texttt{\{\href{mailto:lukas.ellinger@tum.de}{lukas.ellinger}, miriam.anschuetz\}@tum.de, grohg@cit.tum.de}}
\date{}

\begin{document}
\maketitle
\begin{abstract}
Large Language Models (LLMs) can provide accurate word definitions and explanations for any context. However, the scope of the definition changes for different target groups, like children or language learners. This is especially relevant for homonyms---words with multiple meanings---where oversimplification might risk information loss by omitting key senses, potentially misleading users who trust LLM outputs. We investigate how simplification impacts homonym definition quality across three target groups: Normal, Simple, and ELI5. Using two novel evaluation datasets spanning multiple languages, we test DeepSeek v3, Llama 4 Maverick, Qwen3-30B A3B, GPT-4o mini, and Llama 3.1 8B via LLM-as-Judge and human annotations. Our results show that simplification drastically degrades definition completeness by neglecting polysemy, increasing the risk of misunderstanding. Fine-tuning Llama 3.1 8B with Direct Preference Optimization substantially improves homonym response quality across all prompt types. These findings highlight the need to balance simplicity and completeness in educational NLP to ensure reliable, context-aware definitions for all learners.
\end{abstract}

\section{Introduction}
\begin{figure}[t]
    \includegraphics[width=\linewidth]{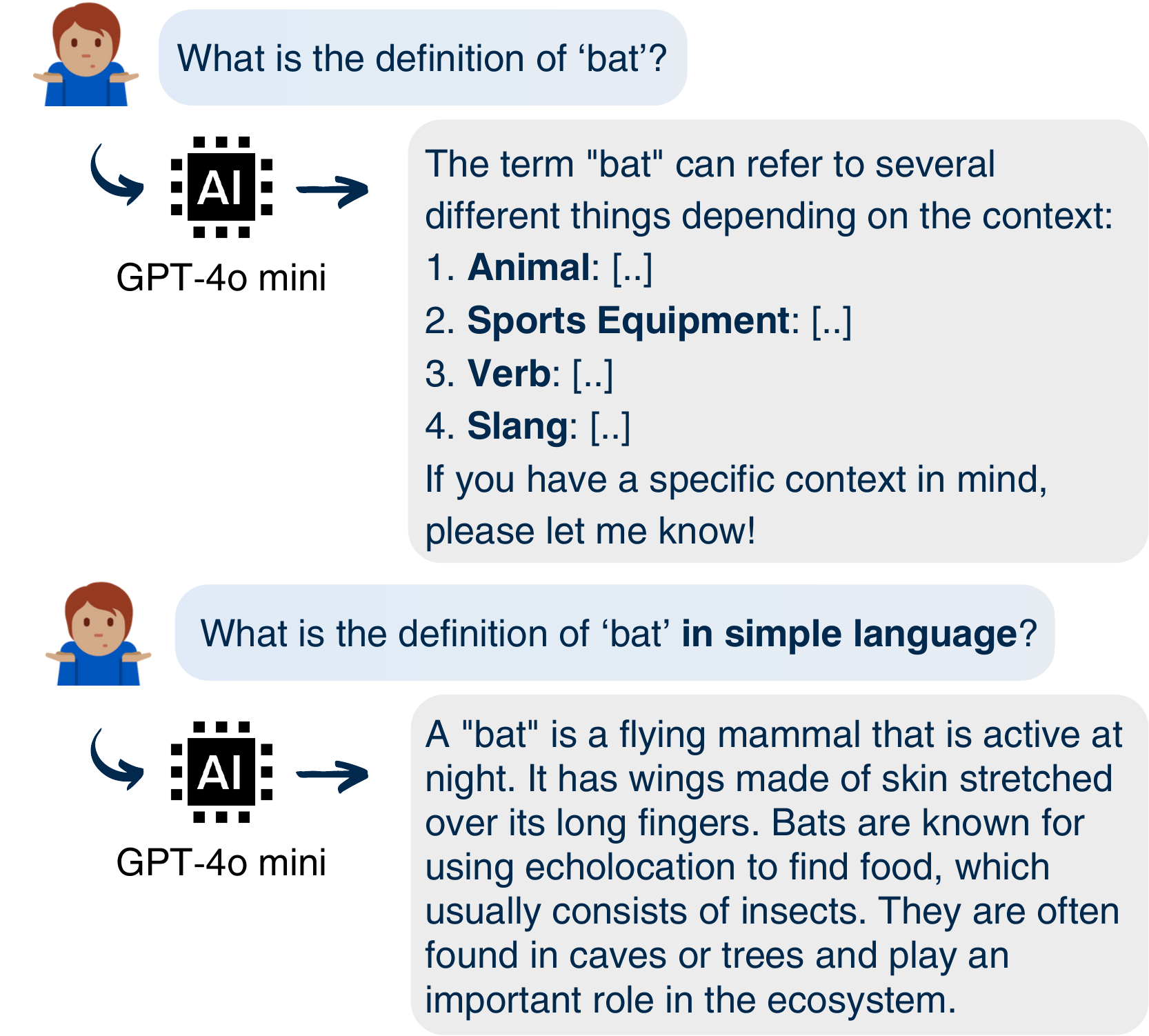}
    \caption{GPT-4o-mini definitions of ``bat'' under normal and simplified style constraints. The normal definition presents multiple senses, showing the word’s ambiguity. The simplified definition provides only one sense, indicating reduced word sense awareness.}
    \label{fig:abstract}
\end{figure}
Large Language Models (LLMs) are increasingly being deployed across diverse domains, with education emerging as an up-and-coming area. These models have the potential to support personalized learning, provide immediate feedback, and increase accessibility for a wide range of learners \citep{xu_large_2024, wang_large_2024, kasneci_chatgpt_2023}.

A core requirement for effective education is personalization. Students vary in their prior knowledge, learning preferences, and cognitive abilities \citep{xu_large_2024}. From young children and university students to individuals with cognitive impairments, learners benefit from tailored content that matches their comprehension level.

However, LLMs can produce false or misleading information \citep{xu_large_2024}. Given their often authoritative tone and rapid accessibility, users, especially those with lower domain knowledge, may uncritically accept incorrect outputs as fact \citep{kasneci_chatgpt_2023}. This is particularly concerning in educational contexts, where reliance on incorrect information can hinder learning and discourage critical thinking.

To improve accessibility, for groups like non-native speakers or individuals with cognitive impairments, educational content is often adapted into simplified, easy-read language \citep{freyer_easy-read_2024}. While this makes information more accessible, it introduces a trade-off: essential information may be lost or oversimplified \citep{trienes_infolossqa_2024}.

One common use case for LLMs is providing instant definitions or explanations across languages. Users often use prompts like ``Explain like I'm five'' (ELI5) to obtain simplified outputs. However, for polysemous words, like homonyms\footnote{We use the term homonym broadly to include both polysemous words and homonyms in the strict linguistic sense (i.e., words with multiple related or unrelated meanings)}, simplification can obscure ambiguity. For instance, Figure \ref{fig:abstract} shows a user requesting definitions for \enquote{bat} in standard and simplified styles from GPT-4o mini. The standard response lists multiple meanings (e.g., animal, sports tool), while the simplified response only mentions the animal. This simplification may mislead users into assuming a single meaning.

This risk is amplified by users’ tendency to assume that LLM responses are both correct and complete. We argue that high-quality definitions should strive for \textbf{completeness}: either (i) enumerate all plausible senses of a homonym, or (ii) exhibit \textbf{Helpful Sense Awareness}. This means clearly stating that not all possible senses are listed, or that additional context is needed for disambiguation. Responses that provide only some senses without such awareness may reinforce misconceptions about the word's meaning.

In this paper, we investigate how well state-of-the-art LLMs handle homonym definition tasks across varying levels of language complexity: Normal, Simple, and ELI5. We assess whether simplification impairs the model's ability to acknowledge ambiguity and provide complete information. Our contributions are as follows:
\begin{itemize}
    \item We propose \textbf{Helpful Sense Awareness}, a novel metric for assessing whether LLMs appropriately acknowledge multiple senses of homonyms during definition tasks.
    \item We present two datasets for evaluating LLM performance on homonym definitions in both multilingual and English settings.
    \item We fine-tune Llama 3.1 8B using Direct Preference Optimization (DPO), greatly improving response quality on homonym definitions.
    \item We empirically demonstrate, using both LLM-as-a-Judge and human annotations, that stylistic constraints aimed at simplification drastically degrade homonym definition quality in models such as DeepSeek-v3, Llama 4 Maverick, Qwen3-30B A3B, GPT-4o mini, and Llama 3.1 8B.
\end{itemize}

\section{Background and Related Work}
\textbf{Definition Modeling}
\citet{noraset_definition_2017} introduced the task of Definition Modeling. The goal is to generate a definition for a given word based on its embedding. Definitions fall into two categories: static and context-dependent. Static definitions are found in dictionaries and lexicons. They provide fixed meanings for words based on predefined word senses, treating each word as having a discrete set of definitions. A related task is Word Sense Disambiguation (WSD). This focuses on identifying the correct sense of a word in a given context by assigning a word sense label \citep{navigli_word_2009}. However, \citet{kilgarriff_i_1997} challenges the idea that word senses are fixed, arguing that meaning emerges from usage and context. This underscores the limitations underlying traditional lexicon-based approaches.

Recent research has predominantly focused on generating context-sensitive definitions, leveraging contextual information to produce precise and relevant meanings \citep{mickus_semeval-2024_2024, periti_automatically_2024, huang_definition_2021, bevilacqua_generationary_2020, ishiwatari_learning_2019, gadetsky_conditional_2018}. \citet{kong_multitasking_2022} focus on generating context-sensitive definition generation in simple language.

\citet{proietti_analyzing_2024} focus on sense selection rather than definition generation. They group WordNet senses based on homonymy relations, mapping them to coarse-grained sense clusters. They then probe contextual representations from pretrained language models (e.g., BERT, DeBERTa) using distance-based metrics, showing that they can distinguish homonymous senses with 95\% accuracy.

In contrast to the focus on context-sensitive methods, users often query “What is [word]?” without additional context. Building on the insight that PLMs can handle lexical ambiguity inherently, we evaluate how LLMs define homonyms in such cases. This allows us to assess their understanding of lexical word senses, their management of ambiguity, and the usefulness of their responses absent contextual cues.

\textbf{Simple Language}
Simplified language enhances accessibility for diverse audiences, including non-native speakers, domain novices, children, and individuals with cognitive impairments. Standards like the Web Content Accessibility Guidelines (WCAG) advocate for its use to foster inclusive communication \citep{w3c_web_2025}. This approach employs straightforward vocabulary, minimal jargon, clear sentence structures, and avoids complex grammar \citep{freyer_easy-read_2024}. It is widely applied in fields such as law, healthcare, and education \citep{garimella_text_2022, deilen_towards_2024, rets_approaches_2022}. Prior studies highlight that simplification in LLM-generated text can introduce omissions or vagueness \citep{trienes_infolossqa_2024, agrawal_text_2024, devaraj_evaluating_2022}. A related approach, Explain Like I'm Five (ELI5), popularized through a dataset of 270,000 Reddit threads \citep{fan_eli5_2019}, simplifies complex topics using analogies and non-technical language. While ELI5 is effective for broad audiences, its impact on preserving content completeness is underexplored.
This paper examines how simplification constraints, including ELI5-style rewriting, affect the completeness of definitions for homonyms, where precise disambiguation is critical.

\textbf{LLM-as-a-Judge} 
LLMs are increasingly used as automatic evaluators across diverse NLP tasks. Prior work has applied models like GPT-3.5 and GPT-4 for evaluation in summarization, dialogue tasks \citep{liu_g-eval_2023}, as well as in fact verification \citep{min_factscore_2023, luo_chatgpt_2023, tang_understanding_2023}.
Employing LLMs as evaluators can drastically reduce annotation efforts while maintaining high evaluation quality. We follow this approach by using GPT-4o-mini to assess response quality in terms of completeness, as detailed in Section \ref{sec:automatic-evaluation}.

\section{Methodology}
This section outlines the datasets, model and prompt configurations, evaluation procedures, and the fine-tuning approach used to improve model preference alignment.

\subsection{Datasets}
\textbf{Multilingual Word-in-Context (ML-WiC)}
We processed the dataset from \citet{martelli_semeval-2021_2021}, designed for Multilingual and Cross-lingual Word-in-Context Disambiguation (MCL-WiC). This task involves determining whether a target word retains the same sense in two sentences within and across languages. We filtered the dataset for words with distinct senses in the same language, identifying them thus as homonyms. The filtered dataset covers 308 Arabic, 334 English, 380 French, 330 Russian, and 254 Simplified Chinese target words.

\textbf{Homonyms with WordNet (HoWN)}
The ML-WiC dataset lacks annotations specifying particular word senses. This limits its utility for evaluating which senses LLMs return in definitions. To address this, we created the Homonyms with WordNet (HoWN) dataset, annotated with senses from WordNet, a comprehensive English lexical database organizing words into synsets representing distinct concepts \citep{miller_wordnet_1994}. To ensure true homonyms with distinct senses, we used coarse-grained sense clusters from \citet{proietti_analyzing_2024}, avoiding overly fine-grained nuances. Using the SemCor Corpus \citep{miller_semantic_1993}, we kept only those that appeared at least twice with different sense annotations. This process yielded 164 target homonyms. Detailed construction steps are provided in Appendix~\ref{app:hown_construction}.

\subsection{Model and Prompt Configuration}
We evaluate five LLMs to assess how style constraints impact the quality of word definition generation: DeepSeek v3 \citep{deepseek-ai_deepseek-v3_2025}, Llama 4 Maverick \citep{meta_ai_llama_2025}, Qwen3-30B A3B \citep{qwen_team_qwen3_2025}, GPT-4o mini \citep{openai_gpt-4_2024}, and Llama 3.1 8B \citep{grattafiori_llama_2024}. These models vary in size, architecture (including Mixture-of-Experts and dense models), and openness, enabling a comprehensive analysis of performance across diverse LLMs. Detailed model specifications are provided in Appendix \ref{app:model_details}.

We use four prompt types to generate definitions:
\begin{itemize}
    \item \textbf{Normal}: ``What is the definition of `\textit{word}'?''
    \item \textbf{Simple}: ``What is the definition of `\textit{word}' in simple language?''
    \item \textbf{ELI5}: ``Explain `\textit{word}' like I am 5 years old.''
    \item \textbf{Multi-Sense-Aware}: Any prompt appended with: ``Keep in mind that some words have more than one meaning.''
\end{itemize}

ELI5 targets explanations for children, Simple seeks definitions in plain language, and Normal allows unconstrained responses. The Multi-Sense-Aware prompt tests whether explicit instructions to account for multiple meanings mitigate stylistic constraints on definition completeness. We list the used prompt types across all languages in Appendix Table \ref{fig:language-prompts}.

\subsection{Response Categorization}
\label{sec:response-categorization}
Ambiguity in language, particularly for homonyms, poses a challenge in natural language processing, as a single word can have multiple meanings depending on context. To evaluate model responses for handling ambiguity, we developed a categorization framework based on three criteria:

\begin{enumerate}
\item \textbf{Number of definitions}: The count of distinct meanings provided for a word.
\item \textbf{Context clarification request}: Whether the response seeks the intended context (e.g., ``Please let me know if you have a specific context in mind!'').
\item \textbf{Remark on additional definitions}: Whether the response acknowledges other unlisted meanings (e.g., ``Here are some of the primary definitions of the word:'').
\end{enumerate}

We define a response as \textbf{complete} if it either provides \textit{all} meanings of a word or exhibits \textbf{Helpful Sense Awareness (HeSA)}.
A response exhibits HeSA by including a \textit{context clarification request} or a \textit{remark on additional definitions}. Such responses are considered high-quality, as they proactively mitigate ambiguity without requiring an exhaustive list of meanings.

A relaxed concept of \textbf{Completeness} is \textbf{Sense Awareness}, which applies when a response includes multiple definitions or demonstrates HeSA. We use Sense Awareness when no dictionary data is available to verify whether extracted definitions match all possible meanings. 
This metric is a valuable simplification of Completeness, as its opposite, providing only one definition without HeSA, is the least desirable outcome for homonyms.

These classifications are crucial because our prompts intentionally lack context. This increases the risk of incomplete responses and misleading users by suggesting an incorrect word meaning. An example of a complete response is provided in Appendix Figure~\ref{fig:response-example}.

\subsection{Automatic Evaluation}
\label{sec:automatic-evaluation}
We designed an automated evaluation framework to categorize model responses using GPT-4o mini as an LLM judge, selected for its efficiency and performance. The framework evaluates responses based on three dimensions outlined in Section~\ref{sec:response-categorization} and extracts all explicitly mentioned definitions. A few-shot prompt, detailed in Appendix~\ref{app:automatic-evaluation}, guides the evaluation. To validate the framework, one author labeled 450 responses from HoWN, with 150 responses per prompt type. The LLM judge achieved 93.33\% agreement on Definition Count Classification (Single, Multiple) and 86.44\% agreement on HeSA compared to these human labels. Further details are provided in Appendix~\ref{app:human-eval}

\subsection{Definition Matching for HoWN}
To analyze sense coverage and preference in HoWN, we mapped the extracted definitions to WordNet senses. We employed a sentence transformer model\footnote{sentence-transformers/all-MiniLM-L6-v2}\citep{reimers_sentence-bert_2019} to compute the cosine similarity between model-generated definitions and the glosses of corresponding WordNet senses. For each generated definition, we selected the sense with the highest similarity score, considering only matches with a minimum similarity of 0.4. Since WordNet senses are ranked by estimated frequency of use \citep{miller_wordnet_1994}, this mapping allowed us to assess which senses the model most frequently aligns with and whether it covers the full range of senses.

\subsection{Direct Preference Optimization}
To improve the completeness of homonym definitions, we fine-tuned the Llama 3.1 8B using DPO \citep{rafailov_direct_2023}. DPO aligns model outputs with desired behavior by training on preference pairs. In our case, we favor complete responses over incomplete ones. We constructed a training set from the HoWN dataset with simple prompts, comparing Llama 3.1 8B's responses to more complete responses from other models to create 116 preference pairs across 63 words. 

We chose HoWN because it allows us to evaluate whether a response captures all possible definitions of a word. Thus, it ensures that completeness is not solely driven by the presence of HeSA, but also by covering all possible senses. Due to the limited data, we did not create a separate validation set. We performed a single training run using the whole training set. This decision reflects our aim to demonstrate the feasibility of aligning models to produce more complete definitions, rather than optimizing for peak performance through extensive tuning. Detailed training information is provided in Appendix~\ref{app:dpo}.

\section{Results}
In this section, we analyze the HoWN dataset, focusing on completeness, sense coverage, and sense distribution. Additionally, we present the results from our DPO fine-tuning, followed by the results of the ML-WIC dataset.

\subsection{HoWN}
\label{hown-results}

\begin{table*}[ht]%
\centering%
\small%
\begin{tabular}{@{}l S[table-format=2.2] S[table-format=2.2] S[table-format=2.2] S[table-format=2.2] S[table-format=2.2] S[table-format=2.2] S[table-format=2.2] S[table-format=2.2]@{}}%
\toprule%
\textbf{Model}&\textbf{FKGL}&\textbf{Sense Aware}&\textbf{Multi. Def.}&\textbf{HeSA}&\textbf{Full}&\textbf{Both}&\textbf{Complete}&\textbf{Covered}\\%
\midrule%
\multicolumn{9}{l}{\textbf{Prompt: Normal}} \\%
Llama 3.1 8B&10.51&\textbf{\tablenum{97.39}}&\textbf{\tablenum{97.39}}&67.97&37.25&26.14&79.08&65.93\\%
GPT{-}4o mini&10.70&95.42&95.42&62.09&49.02&32.68&78.43&74.22\\%
Qwen3{-}30B A3B&8.87&93.46&90.20&\textbf{\tablenum{77.78}}&44.44&\textbf{\tablenum{38.56}}&\textbf{\tablenum{83.66}}&70.20\\%
Llama 4 Maverick&10.76&94.77&94.12&60.13&45.10&25.49&79.74&71.70\\%
DeepSeek v3&10.08&92.81&92.16&22.88&\textbf{\tablenum{53.59}}&7.84&68.63&\textbf{\tablenum{77.04}}\\%
\midrule%
\multicolumn{9}{l}{\textbf{Prompt: Simple}} \\%
Llama 3.1 8B&8.35&71.24&70.59&18.95&17.65&5.23&31.37&51.89\\%
GPT{-}4o mini&8.00&61.44&61.44&7.84&16.99&2.61&22.22&52.88\\%
Qwen3{-}30B A3B&7.20&\textbf{\tablenum{83.01}}&\textbf{\tablenum{82.35}}&22.88&\textbf{\tablenum{30.72}}&\textbf{\tablenum{9.15}}&\textbf{\tablenum{44.44}}&\textbf{\tablenum{60.79}}\\%
Llama 4 Maverick&8.52&73.20&71.90&\textbf{\tablenum{26.14}}&19.61&8.50&37.25&55.29\\%
DeepSeek v3&8.77&61.44&61.44&5.23&20.26&0.00&25.49&54.98\\%
\midrule%
\multicolumn{9}{l}{\textbf{Prompt: ELI5}} \\%
Llama 3.1 8B&4.26&13.73&11.11&3.92&1.31&0.65&4.58&33.41\\%
GPT{-}4o mini&5.27&8.50&8.50&0.00&1.31&0.00&1.31&31.79\\%
Qwen3{-}30B A3B&5.11&\textbf{\tablenum{35.95}}&\textbf{\tablenum{33.33}}&\textbf{\tablenum{9.80}}&\textbf{\tablenum{8.50}}&\textbf{\tablenum{2.61}}&\textbf{\tablenum{15.69}}&\textbf{\tablenum{40.08}}\\%
Llama 4 Maverick&4.03&11.11&9.80&2.61&0.65&0.00&3.27&34.99\\%
DeepSeek v3&5.50&11.11&10.46&1.31&0.65&0.00&1.96&31.67\\\bottomrule%
\end{tabular}%
\caption{Performance metrics for models on the HoWN dataset across Normal, Simple, and ELI5 prompt types. Metrics include Flesch–Kincaid grade level (FKGL), percentage of responses classified as \textit{Sense Aware}, \textit{Multiple Definitions}, \textit{HeSA}, \textit{Full} (covering all meanings), \textit{Both} (HeSA and Full), \textit{Complete}, and the average percentage of coarse-grained definitions covered. Best scores for each prompt type are highlighted in \textbf{bold}.}
\label{tab:hown-overview}%
\end{table*}

\begin{figure}[htbp]
    \includegraphics[width=\linewidth]{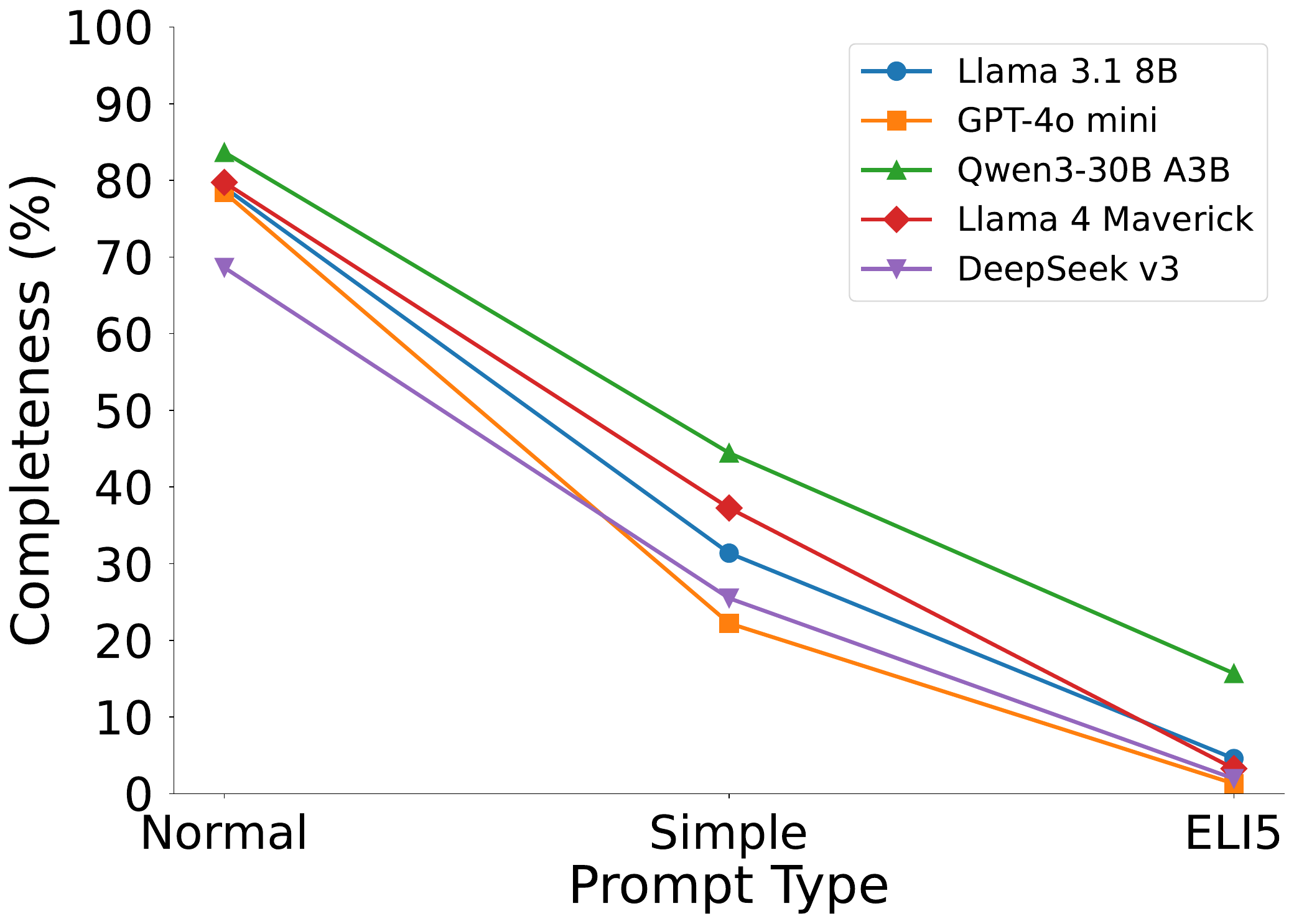}
    \caption{Definition completeness for each model under three stylistic constraints: Normal, Simple, and ELI5}
    \label{fig:completeness-overview}
\end{figure}

To assess the impact of stylistic constraints on model performance, we evaluated response quality using the HoWN dataset across the three prompt types. Figure~\ref{fig:completeness-overview} illustrates the variation in response completeness across these prompt types, highlighting the drastic influence of stylistic constraints.

For the Normal prompt, all models achieve relatively high completeness, ranging from 68.63\% for DeepSeek v3 to 83.66\% for Qwen3-30B A3B. Completeness falls sharply under Simple and ELI5. In the Simple setting, completeness ranges from 22.22\% for GPT-4o mini to 37.25\% for Llama 4 Maverick, while in the ELI5 setting, it ranges from 1.31\% for GPT-4o mini to 15.69\% for Qwen3-30B A3B. Among the models, Qwen3-30B A3B exhibits the smallest drop in completeness across prompt types, while GPT-4o mini shows the largest.

We detail these findings in Table~\ref{tab:hown-overview}. These reveal that simpler prompt types (Simple and ELI5) lead to fewer covered definitions and lower HeSA scores. This suggests that stylistic simplification not only reduces the number of definitions generated but also impairs models’ ability to acknowledge multiple word senses. We further report Sense Awareness and the percentage of responses containing multiple definitions. Both metrics show a marked decline from Normal to Simple to ELI5 prompts, reinforcing the trend. Additionally, we calculate the Flesch-Kincaid grade level (FKGL) \citep{kincaid_derivation_1975} to measure the readability of the generated definition. ELI5 responses have the lowest FKGL (4.03–5.27), followed by Simple (7.20–9.37), and Normal (8.87–10.76). A lower value indicates simpler language, confirming that the models adhere to the simplicity constraints.

\subsubsection{Sense Coverage and Distribution Analysis}
\begin{figure*}[htbp]
    \centering
    \begin{minipage}{0.32\textwidth}
        \centering
        \includegraphics[width=\linewidth]{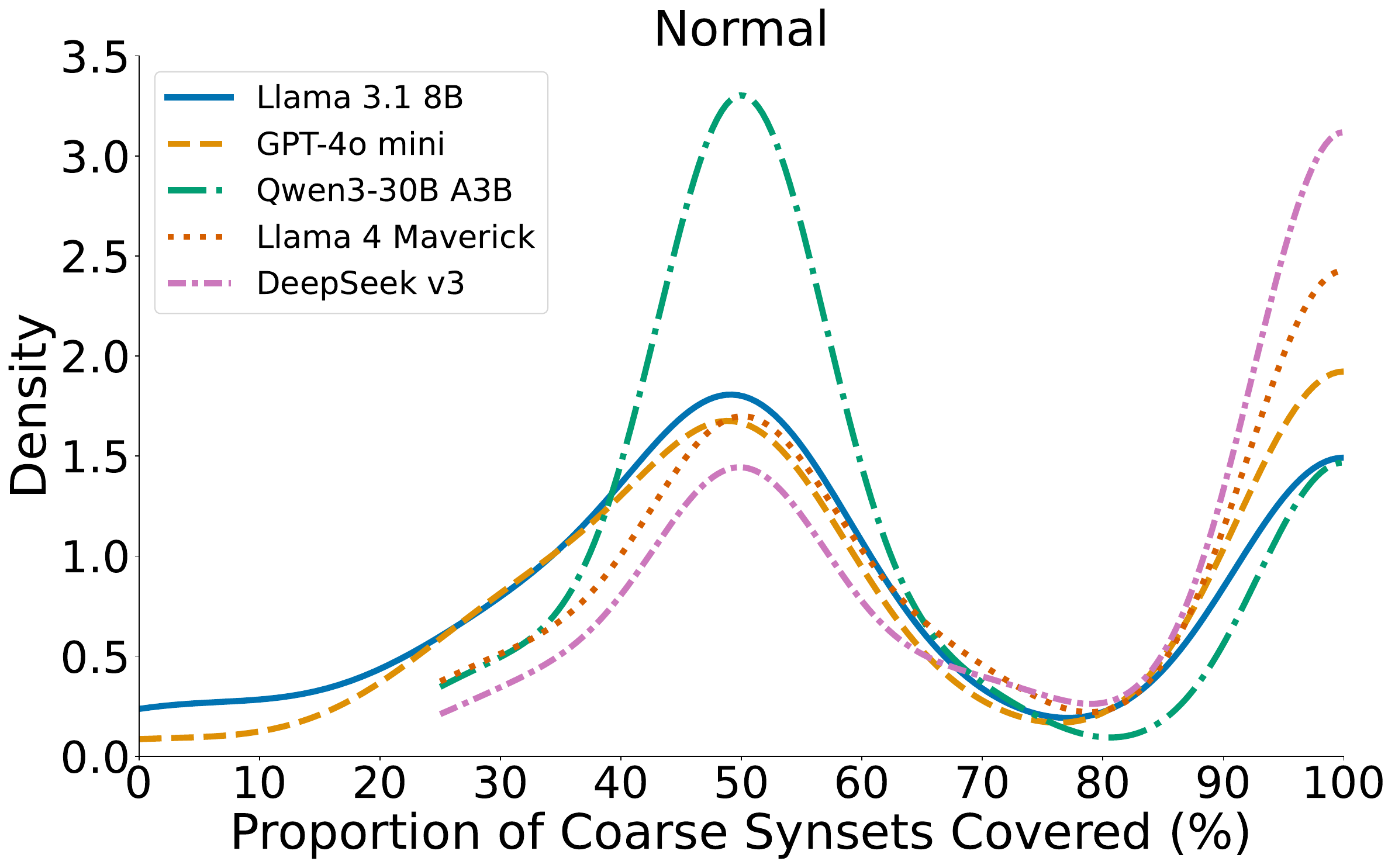}
    \end{minipage}%
    \hfill
    \begin{minipage}{0.32\textwidth}
        \centering
        \includegraphics[width=\linewidth]{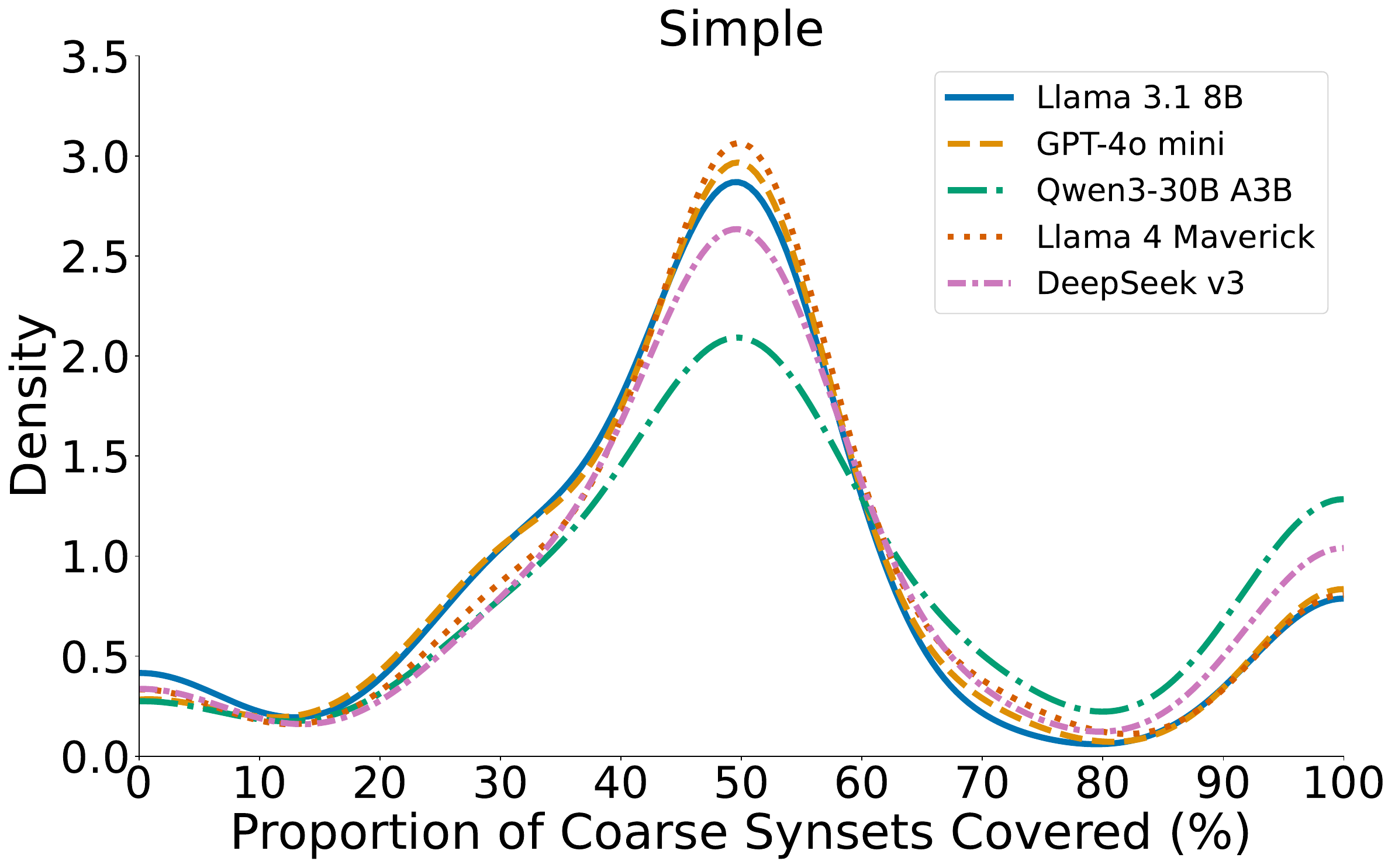}
    \end{minipage}%
    \hfill
    \begin{minipage}{0.32\textwidth}
        \centering
        \includegraphics[width=\linewidth]{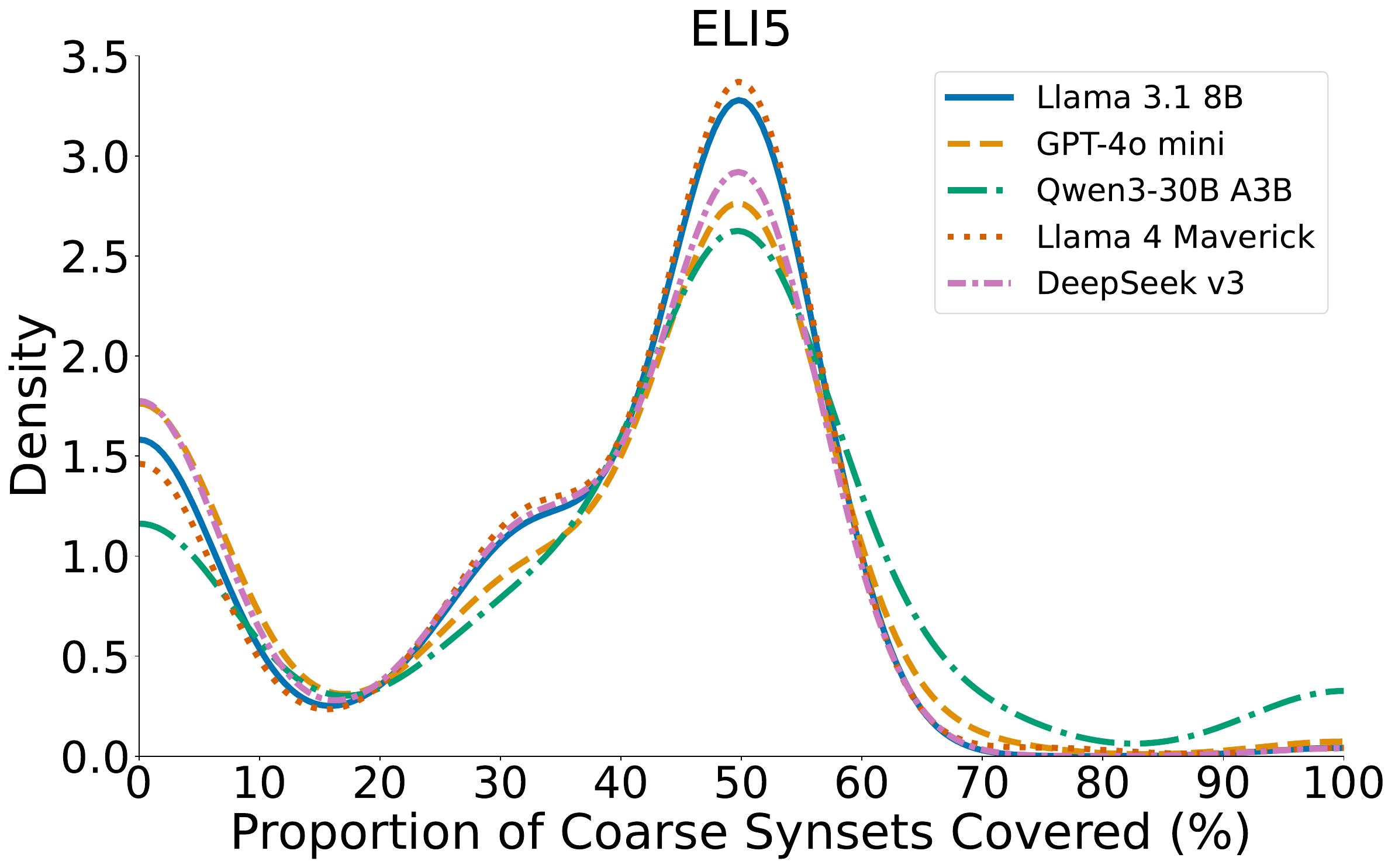}
    \end{minipage}
    \caption{Distribution of coarse sense coverage across model outputs. The x-axis shows synsets coverage, while the y-axis shows estimated density. Only incomplete responses were considered.}
    \label{fig:coarse-senses}
\end{figure*}

As discussed in Section~\ref{sec:response-categorization}, incomplete responses are less desirable due to their potential to mislead users. To evaluate their impact, we analyze the coverage of coarse-grained WordNet senses, as defined by \citet{proietti_analyzing_2024}, for these responses. Figure~\ref{fig:coarse-senses} shows that all prompt styles exhibit a high density of sense coverage around 50\%. For the Normal prompt, there is an additional pronounced peak at 100\% coverage, indicating that many responses capture the full range of senses. The Simple prompt also displays a peak at 100\% coverage, though it is less prominent, with greater density in lower coverage ranges compared to the Normal prompt. In contrast, ELI5 shows less density at higher coverage levels and a greater concentration in lower ranges, suggesting that ELI5-style responses tend to cover fewer senses comprehensively.

Additionally, we analyzed which definitions are most likely to appear in model responses when not all coarse synsets are covered. As expected, more prominent senses tend to be included more frequently than less common ones across all models and prompt types. We provide a detailed breakdown in Appendix Figure~\ref{fig:coarse-senses-coverage}.

\subsubsection{Multi-Sense-Aware Analysis}
We evaluate the impact of prompt types on model performance in a multi-sense-aware setting. We focus on completeness and average synset coverage. Results are presented in Table~\ref{tab:multi-sense-aware-main}.

Context-aware prompting consistently improves completeness across all prompt types, with larger gains for Simple and ELI5 compared to Normal. For instance, Llama 4 Maverick shows substantial improvements: +50.68 in ELI5, +39.19 in Simple, and +15.54 in Normal. Synset coverage likewise increases across all models and prompt types, with stronger gains in Simple and ELI5 settings.  

Interestingly, despite these improvements, completeness still declines from Normal to Simple to ELI5. However, the multi-sense-aware setting substantially mitigates this drop compared to the non-multi-sense-aware baseline. Notably, GPT-4o-mini exhibits a unique decline in completeness (-16.89) under the Normal prompt, indicating model-specific challenges in this setting. We provide detailed results in Appendix~\ref{app:multi-sense-aware}.

\begin{table}[tb]
\centering
\small
\begin{tabular}{@{}l
                S[table-format=2.2]@{\hspace{2pt}}S[table-format=-2.2]
                S[table-format=2.2]@{\hspace{2pt}}S[table-format=-2.2]@{}}
\toprule
\textbf{Prompt / Model} & \multicolumn{2}{c}{\textbf{Complete}} & \multicolumn{2}{c}{\textbf{Covered}} \\
\midrule
\multicolumn{5}{l}{\textbf{Prompt: Normal}} \\
Llama 3.1 8B        & 83.78 & \textcolor{own_green}{\tablenum{+4.05}}  & 73.97 & \textit{\textcolor{own_green}{\tablenum{+8.18}}} \\
GPT-4o mini         & 63.51 & \textcolor{own_red}{\tablenum{-16.89}}   & 78.34 & \textcolor{own_green}{\tablenum{+3.64}} \\
Qwen3-30B A3B       & 91.22 & \textcolor{own_green}{\tablenum{+5.41}}  & 78.47 & \textcolor{own_green}{\tablenum{+7.93}} \\
Llama 4 Maverick    & \textbf{95.95} & \textit{\textcolor{own_green}{\tablenum{+15.54}}} & 77.72 & \textcolor{own_green}{\tablenum{+5.28}} \\
DeepSeek v3         & 76.35 & \textcolor{own_green}{\tablenum{+6.76}}  & \textbf{84.72} & \textcolor{own_green}{\tablenum{+7.45}} \\
\midrule
\multicolumn{5}{l}{\textbf{Prompt: Simple}} \\
Llama 3.1 8B        & 41.89 & \textcolor{own_green}{\tablenum{+10.14}} & 60.55 & \textcolor{own_green}{\tablenum{+8.93}} \\
GPT-4o mini         & 57.43 & \textcolor{own_green}{\tablenum{+35.14}} & 66.71 & \textcolor{own_green}{\tablenum{+13.73}} \\
Qwen3-30B A3B       & 66.22 & \textcolor{own_green}{\tablenum{+21.62}} & 72.69 & \textcolor{own_green}{\tablenum{+11.87}} \\
Llama 4 Maverick    & \textbf{77.70} & \textit{\textcolor{own_green}{\tablenum{+39.19}}} & 73.30 & \textcolor{own_green}{\tablenum{+17.83}} \\
DeepSeek v3         & 56.08 & \textcolor{own_green}{\tablenum{+30.41}} & \textbf{73.49} & \textit{\textcolor{own_green}{\tablenum{+18.68}}} \\
\midrule
\multicolumn{5}{l}{\textbf{Prompt: ELI5}} \\
Llama 3.1 8B        & 39.86 & \textcolor{own_green}{\tablenum{+35.14}} & 54.40 & \textcolor{own_green}{\tablenum{+21.21}} \\
GPT-4o mini         & 30.41 & \textcolor{own_green}{\tablenum{+29.05}} & 52.98 & \textcolor{own_green}{\tablenum{+21.13}} \\
Qwen3-30B A3B       & \textbf{59.46} & \textcolor{own_green}{\tablenum{+43.24}} & 62.52 & \textcolor{own_green}{\tablenum{+22.43}} \\
Llama 4 Maverick    & 54.05 & \textit{\textcolor{own_green}{\tablenum{+50.68}}} & 61.44 & \textcolor{own_green}{\tablenum{+26.96}} \\
DeepSeek v3         & 50.00 & \textcolor{own_green}{\tablenum{+47.97}} & \textbf{62.96} & \textit{\textcolor{own_green}{\tablenum{+31.24}}} \\
\bottomrule
\end{tabular}
\caption{Performance metrics for models on the HoWN dataset under the Multi-Sense-Aware setting across Normal, Simple, and ELI5 prompt types. Metrics show the percentage of \textit{Complete} responses and average percentage of coarse-grained definitions covered, with best scores for each prompt type in \textbf{bold}. Deltas to the non-Multi-Sense-Aware setting are in percentage points, with the largest delta for each prompt type in \textit{italic}.}
\label{tab:multi-sense-aware-main}
\end{table}

\subsection{Direct Preference Optimization}
We assess the impact of DPO fine-tuning on the Llama 3.1 8B model, evaluating its ability to generalize to unseen words. 
Table~\ref{tab:dpo} reports performance metrics on the HoWN dataset, focusing on an evaluation subset of 98 out of 164 words (59.76\%) 
that were not included in the DPO training set. This subset was obtained by excluding three words from the 101 words unseen during training, for which we could not obtain a valid judgment or response from Qwen3-30B A3B, the original Llama 3.1 8B, or its fine-tuned variant.

Despite fine-tuning exclusively on responses to Simple prompts, DPO leads to substantial improvements across all metrics and prompt types. completeness improves by 11.22 percentage points for Normal, 25.51 for Simple, and 14.29 for ELI5. HeSA also improves, with gains of 18.37 points for Normal, 36.73 for Simple, and 9.18 for ELI5. Further, the percentage of covered definitions rises by 6.32 for Normal, 3.28 for Simple, and 6.79 for ELI5, indicating that DPO enhances the model's ability to address multiple word senses beyond the training data.

The FKGL decreases by 0.34 for Normal, indicating simpler language, while increasing by 1.04 for Simple and 0.61 for ELI5. It yet remains comparable to other models in each setting. These results demonstrate that DPO fine-tuning enhances completeness, generalizing effectively to unseen words across diverse prompt types.

On the test subset, our model surpasses Qwen3-30B A3B, the best-performing base model across all prompt types, with completeness improvements of +11.23 (90.82 vs. 79.59) for Normal, +25.51 (75.51 vs. 50.00) for Simple, and +14.26 (19.36 vs. 5.10) for ELI5. Trained only on Simple prompts, DPO fine-tuning enables Llama 3.1 8B to outperform larger models, demonstrating its potential to enhance sense-aware definition generation for unseen homonyms in context-free queries.

\begin{table}[tb]
\centering
\small
\setlength{\tabcolsep}{4pt}
\begin{tabular}{
    l
    @{\hspace{6pt}}
    S[table-format=2.2]@{\hspace{2pt}}r  
    S[table-format=2.2]@{\hspace{2pt}}r  
    S[table-format=2.2]@{\hspace{2pt}}r  
}
\toprule
\textbf{Metric} & \multicolumn{2}{c}{\textbf{Normal}} & \multicolumn{2}{c}{\textbf{Simple}} & \multicolumn{2}{c}{\textbf{ELI5}} \\
\midrule
FKGL         & 10.59 & \textcolor{own_red}{-0.34}   & 9.71  & \textcolor{own_green}{+1.04}   & 5.07  & \textcolor{own_green}{+0.61} \\
Sense Aware  & 98.98 & \textcolor{own_green}{+3.06}  & 86.73 & \textcolor{own_green}{+10.20}  & 35.71 & \textcolor{own_green}{+21.43} \\
Multi. Def.  & 98.98 & \textcolor{own_green}{+3.06}  & 86.73 & \textcolor{own_green}{+11.22}  & 32.65 & \textcolor{own_green}{+21.43} \\
HeSA         & 84.69 & \textcolor{own_green}{+18.37} & 67.35 & \textcolor{own_green}{+36.73}  & 13.27 & \textcolor{own_green}{+9.18} \\
Full         & 44.90 & \textcolor{own_green}{+10.20} & 29.59 & \textcolor{own_green}{+2.04}   & 10.20 & \textcolor{own_green}{+8.16} \\
Both         & 38.78 & \textcolor{own_green}{+17.35} & 21.43 & \textcolor{own_green}{+13.27}  & 4.08  & \textcolor{own_green}{+3.06} \\
Complete     & 90.82 & \textcolor{own_green}{+11.22} & 75.51 & \textcolor{own_green}{+25.51}  & 19.39 & \textcolor{own_green}{+14.29} \\
Covered      & 70.97 & \textcolor{own_green}{+6.32}  & 61.29 & \textcolor{own_green}{+3.28}   & 39.97 & \textcolor{own_green}{+6.79} \\
\bottomrule
\end{tabular}
\caption{Performance metrics for DPO-optimized Llama 3.1 8B on unseen words across Normal, Simple, and ELI5 prompts. Metrics include Flesch–Kincaid grade level, Sense Awareness, Multiple Definitions, HeSA, Full responses, Both (Full and HeSA), Completeness, and percentage of coarse-grained definitions covered. Changes represent percentage point differences compared to the non-DPO model.}%
\label{tab:dpo}
\end{table}

\subsection{ML-WIC}

\begin{figure*}[htbp]
    \centering
    \begin{minipage}[t]{0.19\textwidth}
        \centering
        \includegraphics[width=\linewidth]{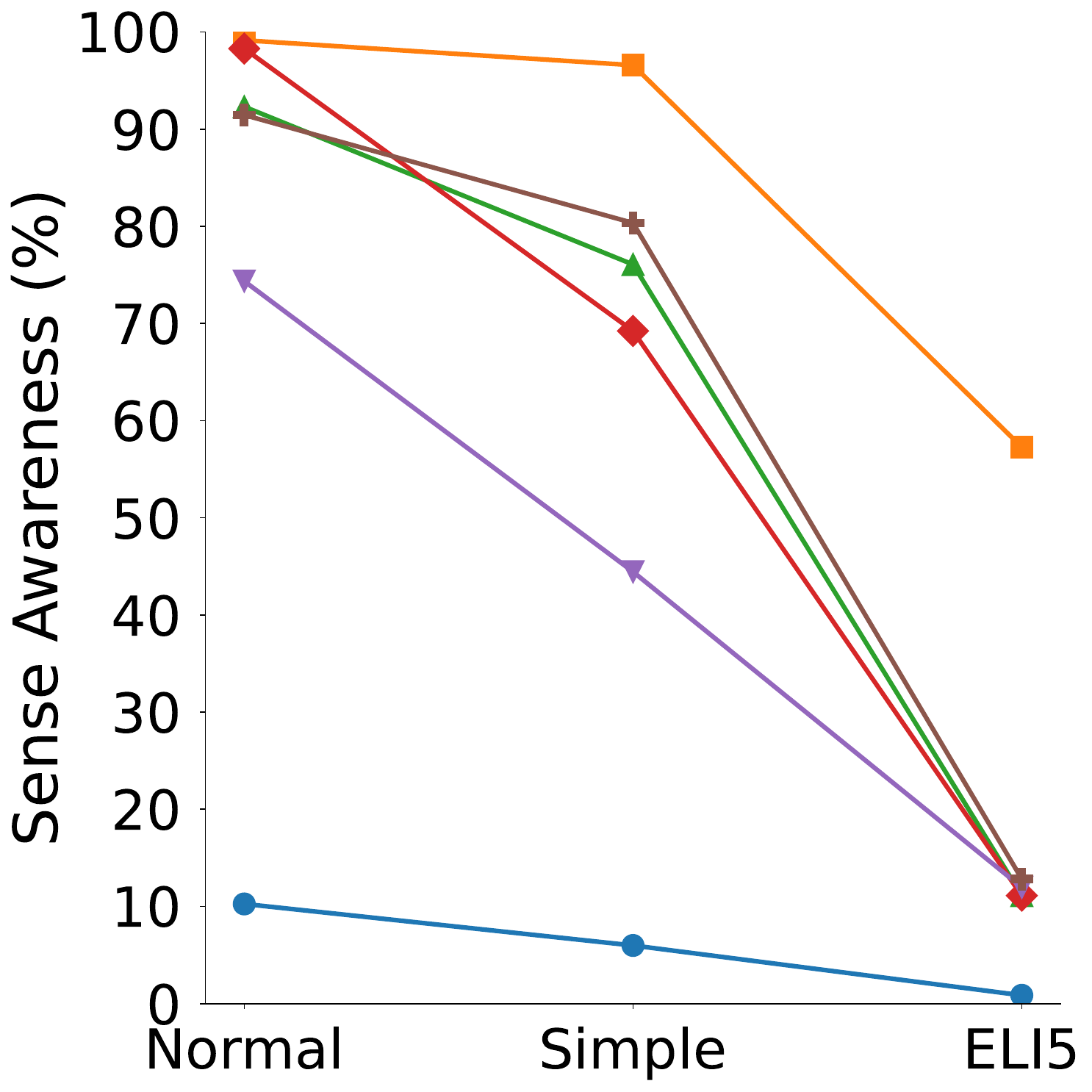}
        \caption*{Arabic}
    \end{minipage}
    \hfill
    \begin{minipage}[t]{0.19\textwidth}
        \centering
        \includegraphics[width=\linewidth]{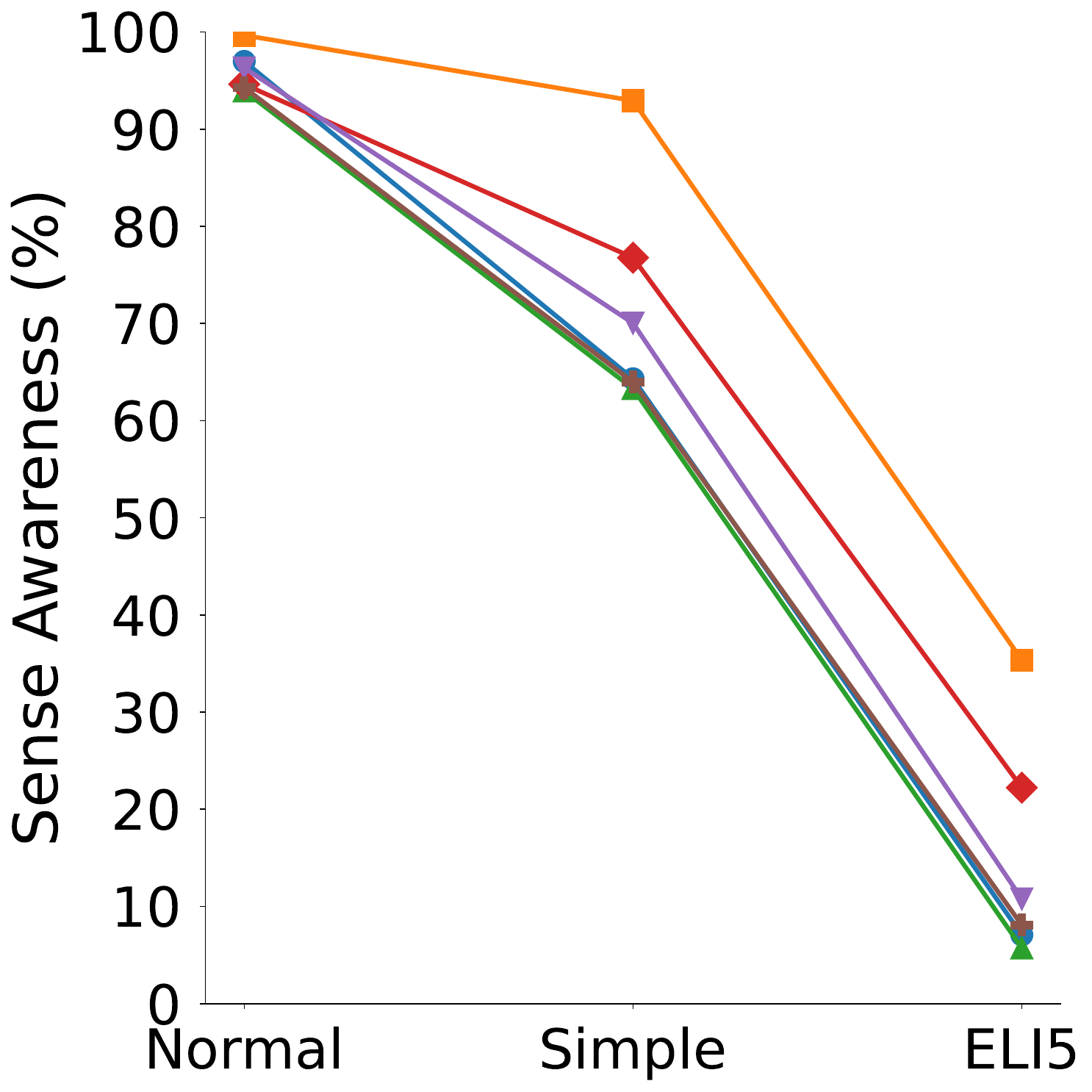}
        \caption*{English}
    \end{minipage}
    \hfill
    \begin{minipage}[t]{0.19\textwidth}
        \centering
        \includegraphics[width=\linewidth]{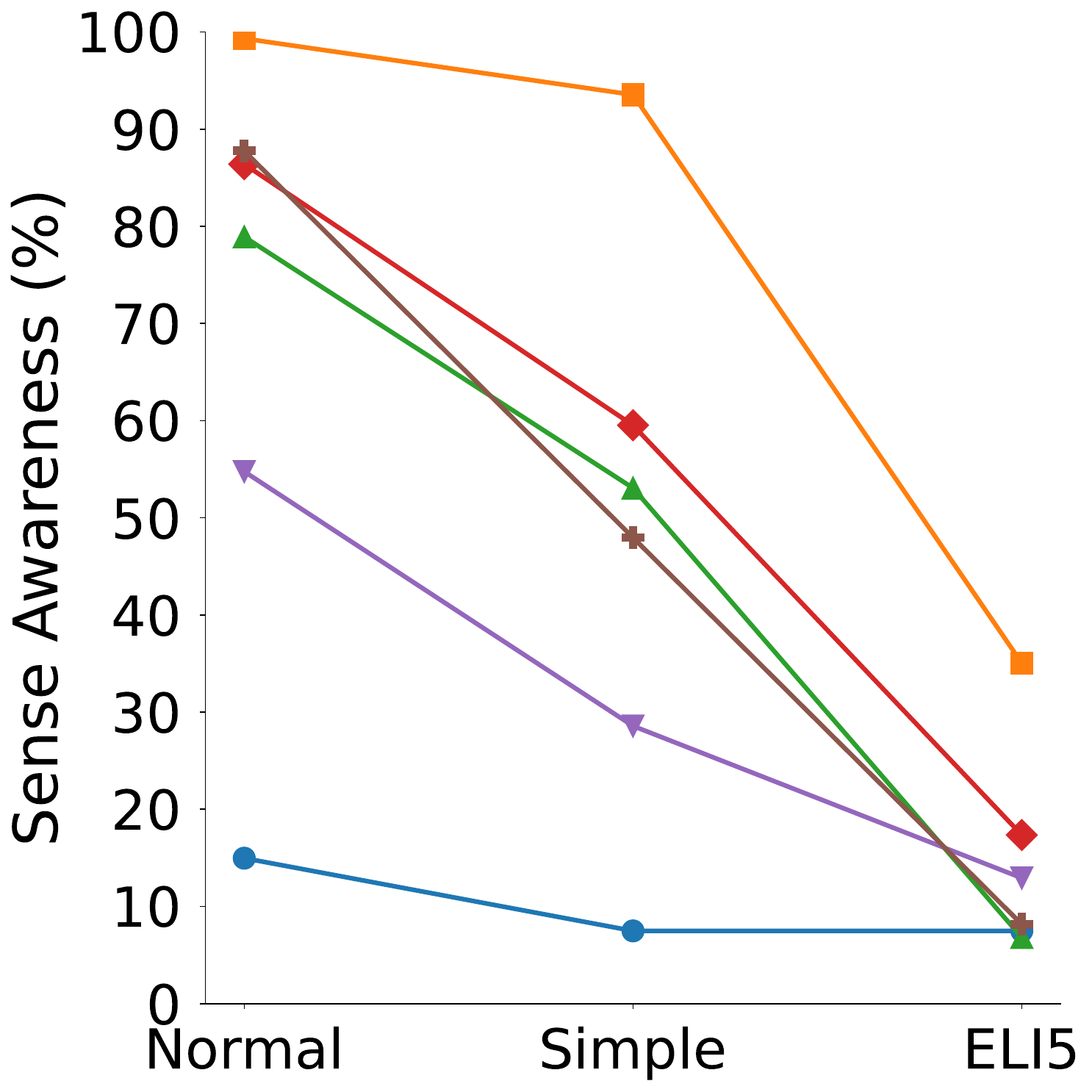}
        \caption*{French}
    \end{minipage}
    \hfill
    \begin{minipage}[t]{0.19\textwidth}
        \centering
        \includegraphics[width=\linewidth]{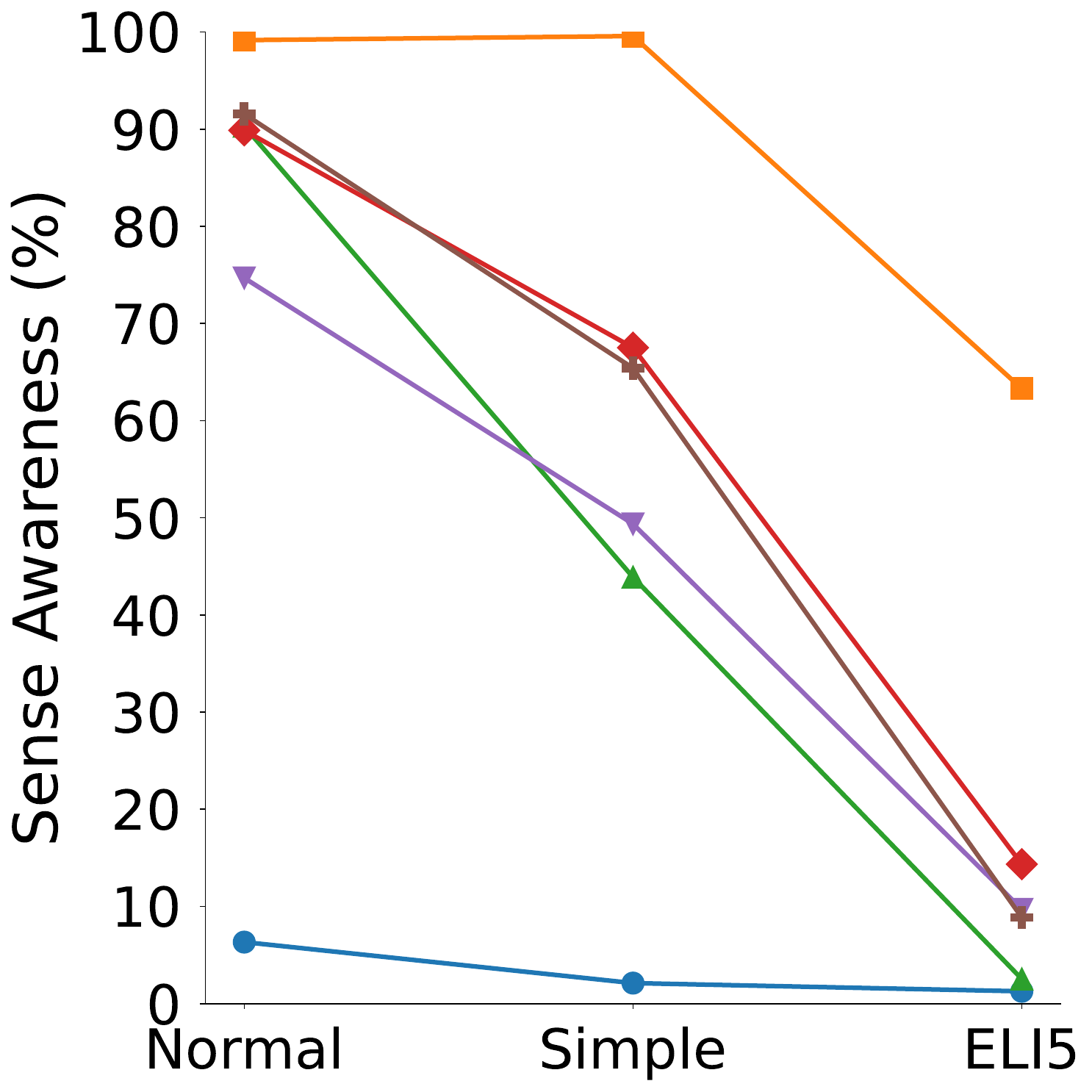}
        \caption*{Russian}
    \end{minipage}
    \hfill
    \begin{minipage}[t]{0.19\textwidth}
        \centering
        \includegraphics[width=\linewidth]{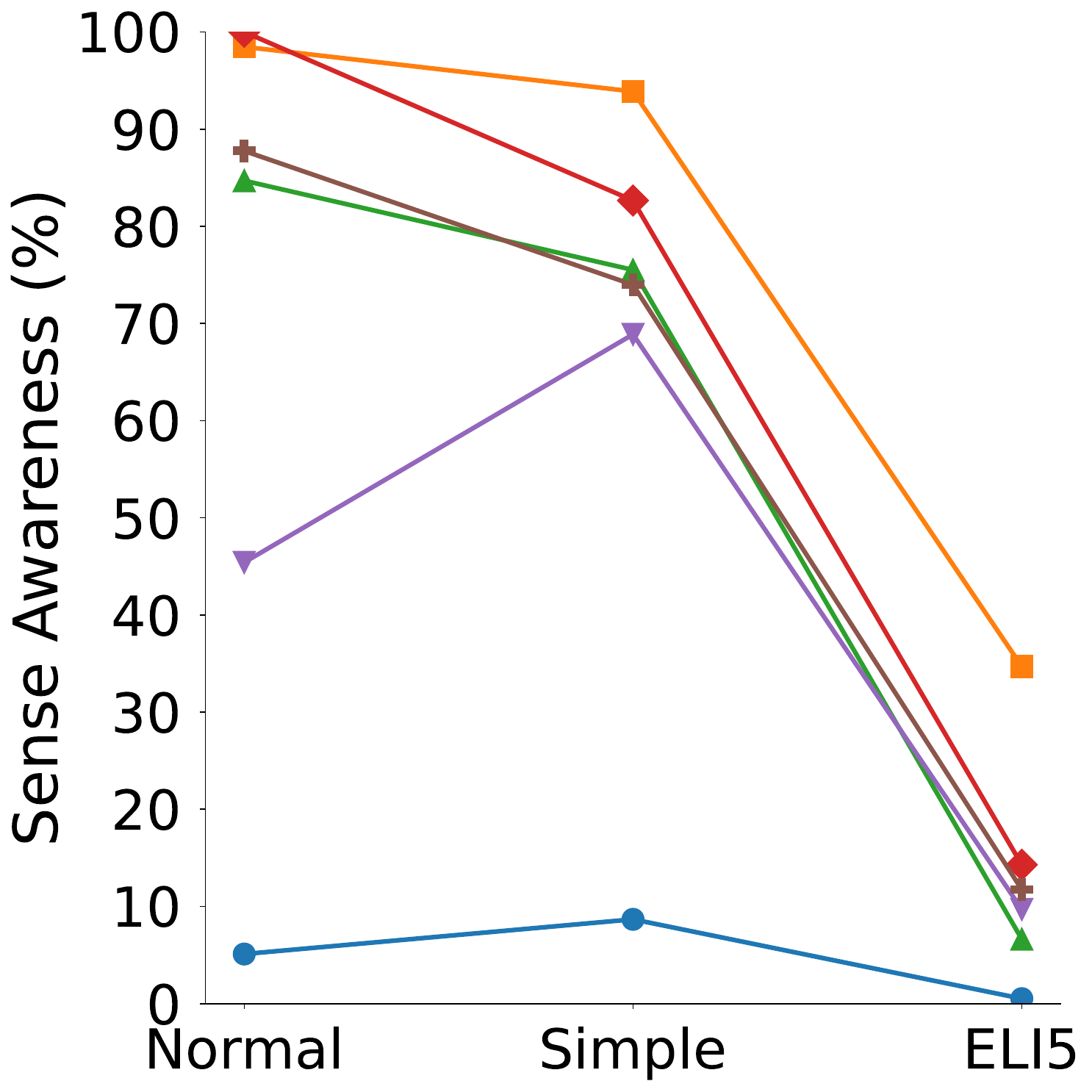}
        \caption*{Chinese (Zh)}
    \end{minipage}
    \caption{Sense Awareness across Languages: Llama 3.1 8B (blue line, circle), DPO Llama 3.1 8B (orange line, square), GPT-4o mini (green line, triangle), Qwen3-30B A3B (red line, diamond), Llama 4 Maverick (purple line, inverted triangle), DeepSeek v3 (brown line, plus).}
    \label{fig:ml-sense-aware}
\end{figure*}

We evaluate prompt type sensitivity across languages using the ML-WiC dataset. Since we do not have dictionary data for all languages, we report Sense Awareness instead of Completeness. Figure~\ref{fig:ml-sense-aware} shows Sense Awareness scores for Arabic, English, French, Russian, and Chinese across the Normal, Simple, and ELI5 prompt types.

We observe a general trend across all languages and models: Sense Awareness declines from Normal to Simple to ELI5. English shows the lowest inter-model variance in Sense Awareness, while also strictly following this trend. Notably, Chinese stands out, as Llama 3.1 8B and Llama 4 Maverick achieve higher Sense Awareness with Simple prompts than with Normal ones, which is an exception to the overall pattern. Among all languages, Chinese also shows the smallest decline from Normal to Simple prompts.

Interestingly, Llama 3.1 8B shows poor Sense Awareness in non-English languages, even with Normal prompts. Our DPO model consistently outperforms all others across prompt types and languages, whereas Llama 3.1 8B, its base model, exhibits the lowest performance. However, except for Russian, our model responded in English despite being prompted in the respective language. We provide a table with more details in Appendix \ref{app:ml-wic}

\section{Discussion}
Our findings reveal that simple language, particularly through Simple and ELI5 prompts, drastically reduces the completeness of homonym definitions. Yet, even Normal prompts yield suboptimal results. In the HoWN dataset, completeness for Normal prompts ranges from 68.63\% (DeepSeek v3) to 83.66\% (Qwen3-30B A3B), falling short of ideal performance (see Table~\ref{tab:hown-overview}). Under ELI5, completeness drops dramatically, falling as low as 1.31\% (GPT-4o mini) to 15.69\% (Qwen3-30B A3B). Similarly, definition coverage is lower in simpler settings, indicating a loss of nuanced meanings.

We observe a similar trend in Sense Awareness across all languages in the ML-WIC dataset. Further, inter-model variance differs across prompt types and languages in the multilingual setting. This variance likely reflects the influence of each language during training. Additionally, Llama 3.1 8B demonstrates notably low performance in non-English languages. Here, its smaller model size probably contributes to this limitation.

The tendency of LLMs to provide only one or a subset of definitions in simpler settings may first seem intuitive, as it reduces complexity. However, this reflects a misunderstanding of simplification. Simplification should enhance understandability without sacrificing information. While it may be reasonable to present fewer senses in simpler settings, responses should still incorporate HeSA to ensure completeness. These findings align with prior work by \citet{trienes_infolossqa_2024}, highlighting simplification-induced information loss in question-answering tasks. Similarly, \citet{kong_multitasking_2022} analyzed context-dependent definitions and found that simplified versions achieve lower BLEU scores and sentence similarity than complex ones, further supporting the trade-off between simplification and information retention.

To address these challenges, we employed DPO and Multi-Sense-Aware Prompting, which both showed promising improvements in definition quality. However, we did not evaluate these methods on single-definition words, where they might have unintended effects, such as overcomplicating responses. An alternative can be Steering Vectors, used by \citet{rimsky_steering_2024}. These can enhance model behavior at inference time without increasing prompt length (as in Multi-Sense-Aware Prompting) or requiring fine-tuning (as in DPO).

In the multilingual setting, the performance of our DPO model reveals intriguing patterns. The base model exhibits by far the weakest performance in non-English languages, whereas the DPO-tuned model consistently outperforms others across all scenarios. Notably, the DPO model frequently responds in English, even when prompted in other languages, which may reflect its training bias. As we did not verify the factual accuracy of responses, some outputs may include hallucinations. Nevertheless, the success of DPO fine-tuning remains evident: Despite training a relatively small model on a limited English dataset in the simple prompt setting, it achieved superior performance across all prompt types and languages.
We thus argue that LLMs are capable of giving complete homonym definitions. However, they are limited by the expected model behavior.

\section{Conclusion}
In this paper, we analyze how LLMs behave in homonym definition generation without additional context, a setting that requires context-independent understanding. We have shown that LLMs have an overall reduced multi-sense awareness, especially for simplified outputs, indicating an oversimplification of contents.

These findings highlight the need for LLMs to better serve diverse users, particularly marginalized groups like non-native speakers and individuals with cognitive impairments. These groups encounter challenges due to linguistic exclusion, underscoring the importance of inclusive design \citep{freyer_easy-read_2024}. Improving LLM inclusivity is essential for building more accessible and effective language technologies.

To support reproducibility and future research, we release our code\footnote{\href{https://github.com/lukasellinger/homonym-eval}{https://github.com/lukasellinger/homonym-eval}}. Further links to models and datasets are provided in the repository.

\section{Limitation}
\textbf{LLM Judge}
The evaluation scores of our LLM judge are sensitive to scoring prompt wording, potentially introducing variability. Inherent LLM biases may also cause systematic differences from human judgments~\cite{rimsky_steering_2024}. Additionally, prior work suggests LLMs underperform in simplified settings, which may affect automated evaluation reliability~\citep{anschutz_simpler_2024, anonymous_disimproved_2025}. Nonetheless, our human evaluation closely mirrors LLM judge scores. The differences observed between prompt types far exceed any potential error margin, strongly reinforcing our findings' robustness.

\textbf{Selected Prompts}
Our study employed three predefined prompts to elicit definitions, reflecting choices a typical user might make without optimizing for prompt variation. However, LLMs are highly sensitive to prompt phrasing, which can impact response quality~\cite{brown_language_2020}. While our approach mirrors realistic user behavior, it does not account for potential gains from prompt optimization. Future research could systematically investigate the effects of varied or optimized prompts on LLM performance.

\textbf{Factuality of Definitions}
We did not verify the factual accuracy of the definitions generated by the models. Consequently, a response may be complete and adhere to the HeSA framework but contain factually incorrect definitions.

\textbf{Reliance on WordNet}
Our analysis of definition completeness and returned definitions relied solely on WordNet. Although WordNet is a widely adopted resource, this choice may not capture definitional nuances in other databases. Future studies could incorporate alternative resources, such as ConceptNet~\cite{speer_conceptnet_2017-1} or Wiktionary, to validate our findings across diverse lexical datasets.

\section*{Acknowledgments}
This research has been funded by the German Federal Ministry of Research, Technology and Space (BMFTR) through grant 01IS23069 Software Campus 3.0 (Technical University of Munich) as part of the Software Campus project \enquote{LIANA}.

\bibliographystyle{acl_natbib}
\bibliography{main}

\appendix
\section{HoWN Construction}
\label{app:hown_construction}
The Homonyms with WordNet (HoWN) dataset was constructed to evaluate LLM performance in generating definitions for homonyms with distinct senses. We filtered for nouns with at least two distinct definitions and more than two characters to exclude short, ambiguous terms. Nouns were chosen to avoid part-of-speech ambiguities; for example, words like \textit{run} can function as both a noun (e.g., a jog) and a verb (e.g., to move quickly), with senses that are often closely related. Focusing on nouns ensured a clear separation of meanings, suitable for homonym definition tasks. This initial filtering yielded 1,780 candidate homonyms. To ensure these homonyms had multiple prevalent senses, rather than a dominant sense with rare secondary ones, we further filtered using the SemCor Corpus \citep{miller_semantic_1993}, the largest manually sense-annotated corpus based on WordNet, containing 226,040 sense annotations across 352 documents. We retained only words with at least two occurrences in SemCor, each annotated with a different sense, confirming their polysemy in real-world usage. This final step resulted in 164 target homonyms, each with verified, distinct senses suitable for evaluating LLM definition generation.

\section{Selected Models}
\label{app:model_details}
We evaluated the following language models, selected for their diversity in size, architecture, and accessibility:

\begin{itemize}
\item \textbf{DeepSeek-V3} \citep{deepseek-ai_deepseek-v3_2025}: An open-weight Mixture-of-Experts (MoE) model with 671 billion total parameters, activating 37 billion per token.

\item \textbf{Llama 4 Maverick}\footnote{\url{https://huggingface.co/meta-llama/Llama-4-Maverick-17B-128E-Instruct}}: A 400-billion-parameter MoE model with 128 experts, activating 17 billion parameters per token.

\item \textbf{Qwen3-30B A3B} \citep{qwen_team_qwen3_2025}: A 30-billion-parameter model from Alibaba's Qwen series, using an MoE architecture. We utilized its thinking mode for enhanced reasoning

\item \textbf{GPT-4o mini}\footnote{gpt-4o-mini-2024-07-18, https://platform.openai.com/docs/models/gpt-4o-mini}: A lightweight variant of OpenAI's GPT-4o, optimized for faster response times and lower resource usage while maintaining competitive performance.

\item \textbf{Llama 3.1 8B} \citep{grattafiori_llama_2024}: An 8-billion-parameter model from Meta's Llama 3.1 series.
\end{itemize}

\section{Response Categorization}
\begin{figure}[htbp]
\includegraphics[width=\linewidth]{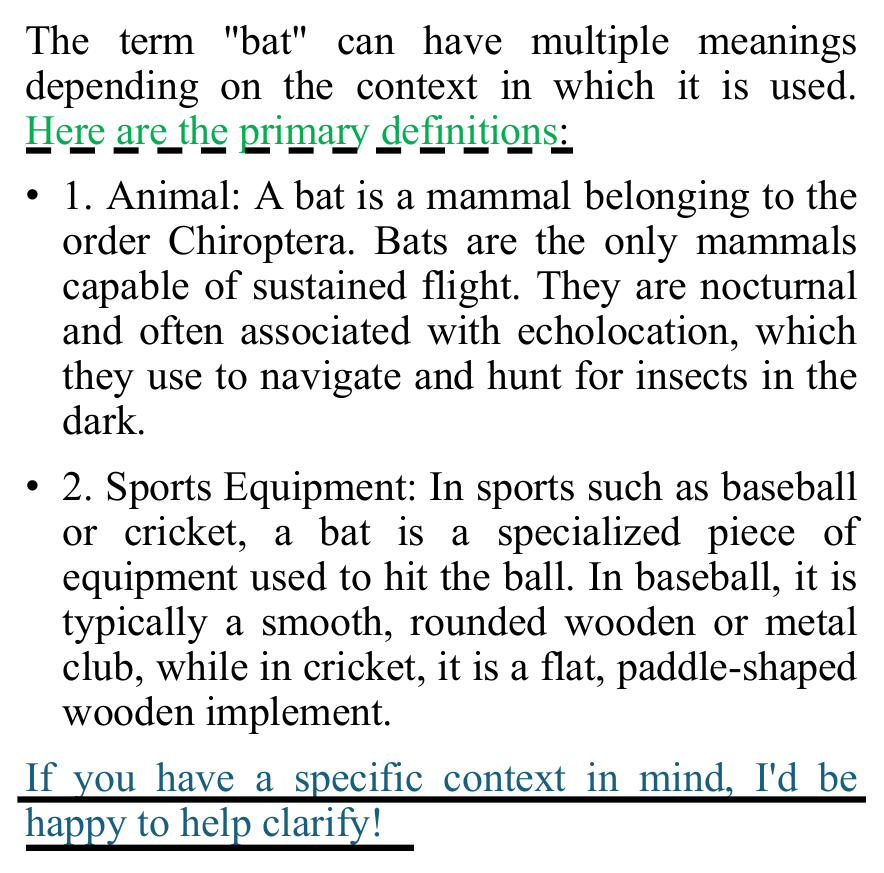}
\caption{Example response (adapted from DeepSeek v3). The green dashed line highlights the HeSA-style remark, while the blue underlined segment shows the context clarification request.}
\label{fig:response-example}
\end{figure}

Figure~\ref{fig:response-example} illustrates an example of a complete response, which includes all definitions (Full) as well as both key characteristics of Helpful Sense Awareness (HeSA): (i) a remark indicating that not all possible meanings may be covered, and (ii) a request for additional context.

\section{Automatic Evaluation}
\label{app:automatic-evaluation}
We found that splitting the categorization into two separate tasks, one for definition extraction and one for Helpful Sense Awareness (HeSA), improves evaluation accuracy. Additionally, we use a slightly adapted version of the definition extraction prompt for the ELI5 setting, as its stylistic characteristics differ from those of the Normal and Simple prompts. Figure~\ref{fig:hesa-prompt} shows the system prompt for HeSA extraction. Figure~\ref{fig:defintion-prompt} provides the system prompt for standard definition extraction. Figure~\ref{fig:eli5-prompt} shows the adapted prompt for definition extraction in the ELI5 setting. Finally, Figure~\ref{fig:user-prompt} presents the user prompt layout used across all tasks.

\section{Direct Preference Optimization}
\label{app:dpo}
We fine-tuned LLaMA 3.1 8B using Direct Preference Optimization (DPO) \citep{rafailov_direct_2023} to improve the completeness of homonym definitions. DPO trains a model on a dataset of prompt-response pairs, where each prompt is associated with a preferred (complete) and a non-preferred (incomplete) output. We constructed our training dataset using the HoWN dataset with the simple prompt type. For each response generated by LLaMA 3.1 8B, we evaluated its completeness. Incomplete responses were labeled as non-preferred, while complete responses from other models for the same prompt were labeled as preferred, yielding 116 preference pairs for 63 words. This process enhanced the model’s ability to generate comprehensive definitions for homonyms. We provide fine-tuning details in Table~\ref{tab:dpo-finetuning-specs}.

\begin{table}[h]
\centering
\begin{tabular}{ll}
\toprule
\textbf{Parameter} & \textbf{Value} \\
\midrule
\multicolumn{2}{l}{\textit{LoRA Configuration}} \\
$r$ & 64 \\
LoRA Alpha & 16 \\
LoRA Dropout & 0.05 \\
Target Modules & \makecell[l]{[q\_proj, v\_proj,\\ k\_proj, o\_proj]} \\
Bias & none \\
\midrule
\multicolumn{2}{l}{\textit{DPO Training Configuration}} \\
$\beta$ & 0.1 \\
Learning Rate & 5e-5 \\
Batch Size (per device) & 4 \\
Epochs & 10 \\
\bottomrule
\end{tabular}
\caption{Combined configuration used for LoRA adaptation and Direct Preference Optimization (DPO) fine-tuning.}
\label{tab:dpo-finetuning-specs}
\end{table}

\section{Human Evaluation}
\label{app:human-eval}
To assess the reliability of our automated evaluation using an LLM-Judge, we conducted a human evaluation
on the HoWN dataset. We randomly sampled 450 responses, with 150 responses per prompt type (ELI5, Simple, and Normal), annotated by one of the authors.
We sampled responses from all models: Llama 4 Maverick (50 ELI5, 50 Normal, 50 Child), DeepSeek v3 (50 Simple),
Qwen3-30B A3B (50 Normal), DPO Llama 3.1 (50 Normal, 50 Simple, 50 Child), Llama 3.1 (50 Simple), and GPT-4o mini (50 Child).
We computed accuracy and Cohen’s Kappa score to measure agreement between the human annotator and the framework for two tasks:
Definition Count Classification (Category) and Helpful Sense Awareness (HeSA).

Table~\ref{tab:human-eval} reports the Cohen’s Kappa scores and accuracy across prompt types.
The combined Kappa scores are 0.86 for Category and 0.71 for HeSA, indicating almost perfect and substantial agreement,
respectively \citep{landis_measurement_1977}. The average accuracy is 93.33\% for Category and 86.44\% for HeSA.

The variability in Kappa scores across individual prompt types stems from their distinct characteristics.
For ELI5, most responses contain a single definition and lack HeSA. Conversely, Normal prompt responses exhibit
a more balanced distribution of definitions and HeSA. This prompt-specific distribution affects the agreement scores,
as Category and HeSA highly depend on the prompt type.

\begin{table}[h!]
\centering
\begin{tabular}{lcc}
\toprule
\textbf{Prompt} & \textbf{Category} & \textbf{HeSA} \\
\midrule
ELI5   & 0.76 (93.33) & 0.17 (94.67) \\
Simple & 0.81 (92.00) & 0.65 (84.67) \\
Normal & 0.57 (94.67) & 0.49 (80.00)  \\
\textbf{Combined} & 0.86 (93.33) & 0.71 (86.44) \\
\bottomrule
\end{tabular}
\caption{Cohen’s Kappa scores and accuracy percentages (in parentheses) for Definition Count Classification (Category) and Helpful Sense Awareness (HeSA) across prompt types on the HoWN dataset.}
\label{tab:human-eval}
\end{table}

\section{Sense Coverage and Distribution Analysis}
\begin{figure*}[htbp]
    \centering
    \begin{minipage}{0.32\textwidth}
        \centering
        \includegraphics[width=\linewidth]{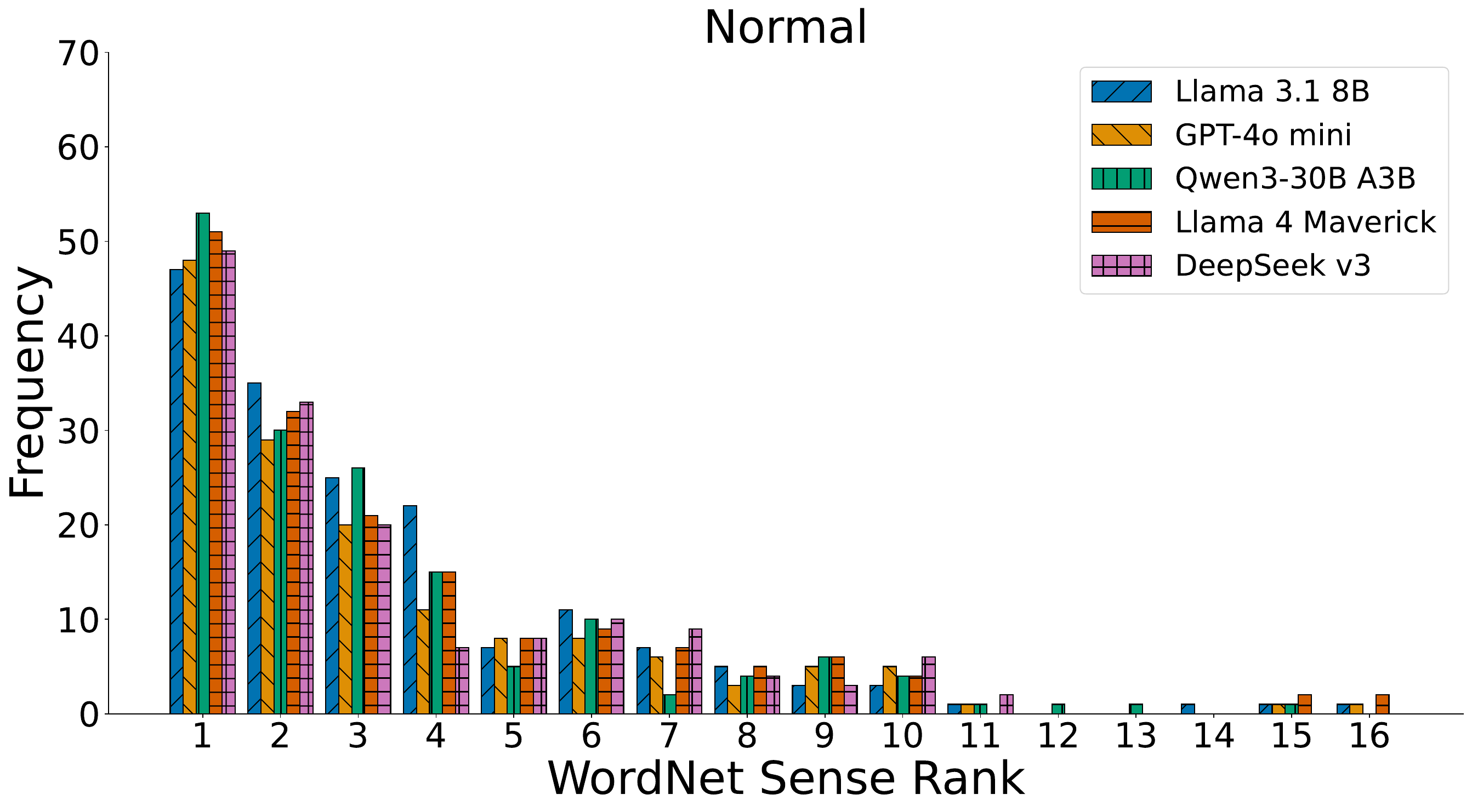}
    \end{minipage}%
    \hfill
    \begin{minipage}{0.32\textwidth}
        \centering
        \includegraphics[width=\linewidth]{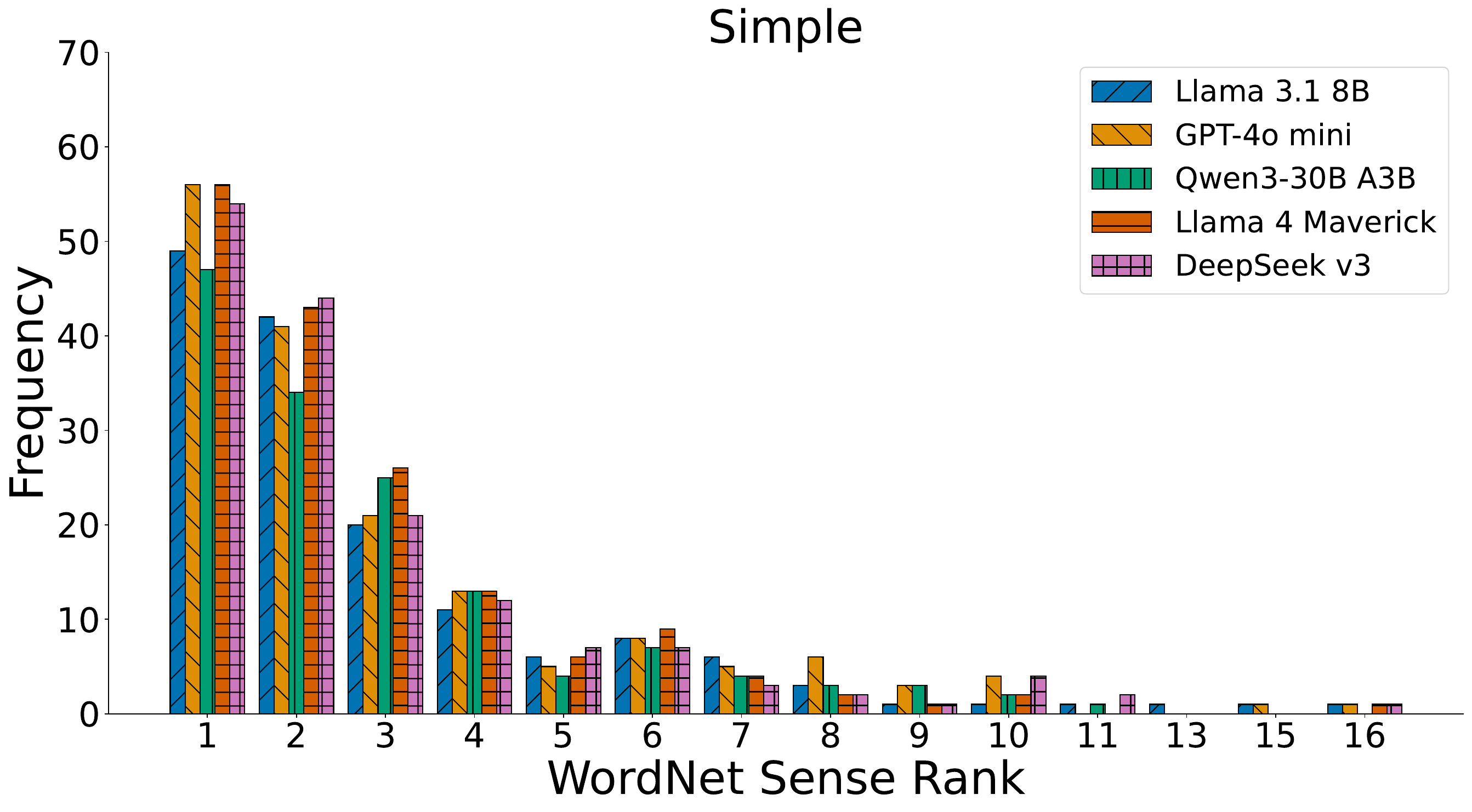}
    \end{minipage}%
    \hfill
    \begin{minipage}{0.32\textwidth}
        \centering
        \includegraphics[width=\linewidth]{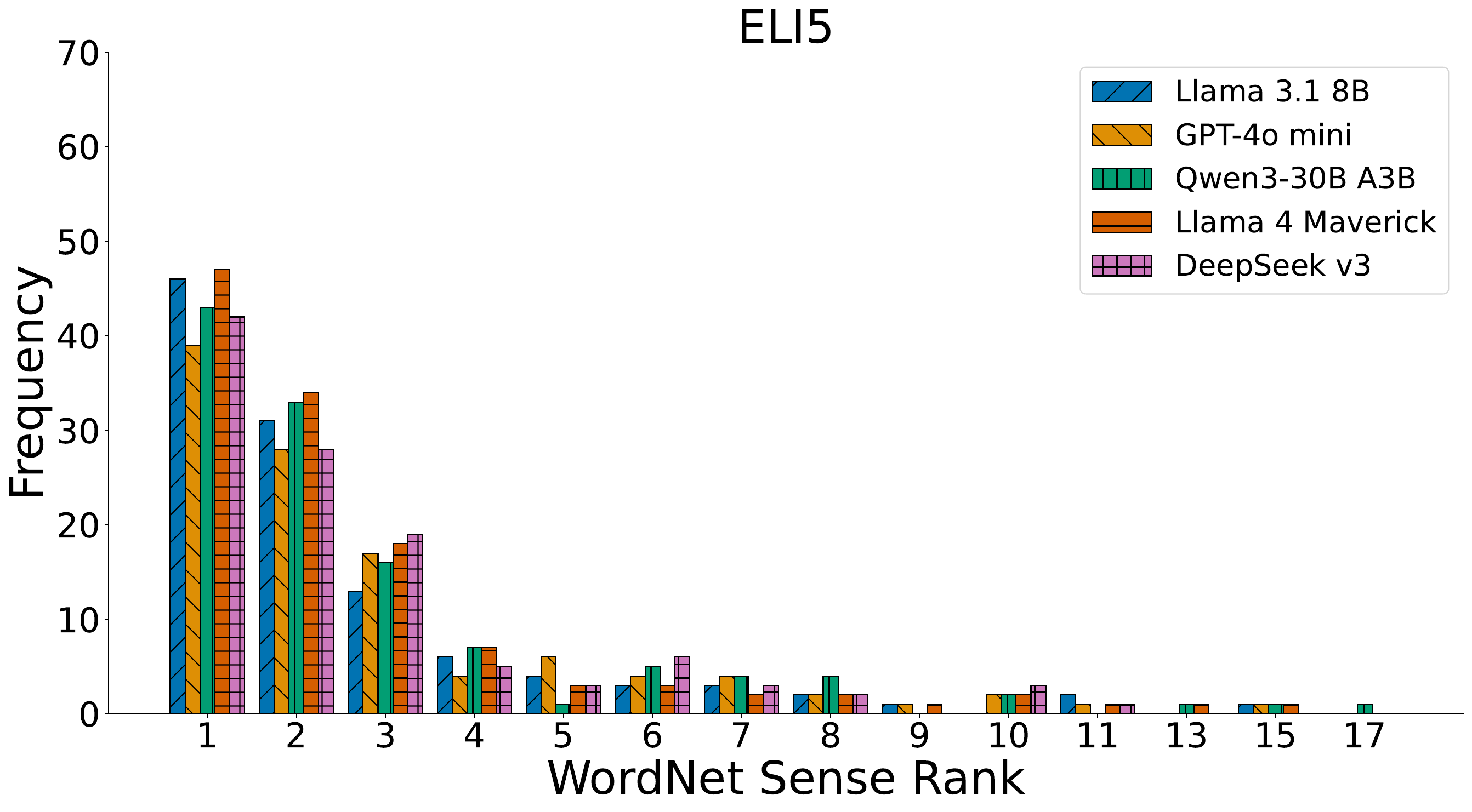}
    \end{minipage}
    \caption{Frequency distribution of covered WordNet sense ranks across model-generated definitions. The x-axis represents the rank of WordNet senses, with lower ranks corresponding to more commonly used senses. The y-axis shows the frequency of occurrences for each rank, reflecting how often the model associates definitions with higher or lower-ranked senses. Only responses with partial coverage of coarse synsets were considered.}
    \label{fig:coarse-senses-coverage}
\end{figure*}

We analyzed which definitions most likely appear in model responses when not all coarse-grained synsets are covered. WordNet sense ranks are ordered by estimated frequency of use, with lower ranks corresponding to more commonly used senses. We matched the extracted model definitions to their corresponding WordNet senses to enable this analysis. As shown in Figure~\ref{fig:coarse-senses-coverage}, more prominent senses are included more frequently than rarer ones across all models and prompt types, which Based on our definition matching, we note that models occasionally describe the same WordNet sense more than once. As shown in Table~\ref{tab:duplicate-wordnet}, this occurs in an average of 56.51\% of multi-definition responses for the Normal prompt, 31.29\% for Simple, and 10.50\% for ELI5. This difference can be attributed to fewer definitions being generated in simpler settings. The Normal prompt tends to be more fine-grained, often presenting the same sense in different contexts. As a result, the frequency of covered word senses remains relatively consistent across prompt types in Figure~\ref{fig:coarse-senses-coverage} since we count only whether a sense is covered, not how often.

\begin{table}[h!]%
\centering%
\setlength{\tabcolsep}{4pt}%
\begin{tabular}{@{}l c c c@{}}%
\toprule%
\textbf{Model}&\textbf{Normal}&\textbf{Simple}&\textbf{ELI5}\\%
\midrule%
Llama 3.1 8B&68.46&50.93&11.76\\%
GPT{-}4o mini&52.74&21.28&0.00\\%
Qwen3{-}30B A3B&62.32&25.40&7.84\\%
Llama 4 Maverick&43.06&35.45&26.67\\%
DeepSeek v3&56.03&23.40&6.25\\\bottomrule%
\end{tabular}%
\caption{Duplicate WordNet Table.}%
\label{tab:duplicate-wordnet}
\end{table}

\section{Multi-Sense-Aware Analysis}
\label{app:multi-sense-aware}
Table~\ref{tab:multi-sense-aware} extends Table~\ref{tab:multi-sense-aware-main} from the main section. Specifically, it includes additional metrics for \textit{FKGL}, \textit{Sense Awareness}, \textit{Multiple Definitions}, \textit{HeSA}, \textit{Full}, and \textit{Both}.

In the Normal setting, all models demonstrate perfect Sense Awareness by providing multiple definitions in every response, except for Qwen3-30B A3B, which falls just short with 99.32\%. A similar pattern appears in the Simple setting, where Sense Awareness remains nearly perfect. GPT-4o mini achieves the lowest score at 95.27\%, which is also the lowest among all models that provide multiple definitions. In the ELI5 setting, Sense Awareness also increases, with GPT-4o mini again at the lower end, reaching 85\% Sense Awareness and 84.46\% for multiple definitions. This aligns with expectations, as the system prompt explicitly encourages comprehensive explanations through the instruction:``Keep in mind that some words have more than one meaning.'' HeSA also improves significantly for the Simple and ELI5 prompts. For the Normal prompt, however, the impact varies by model. GPT-4o mini shows a substantial decline in HeSA, with a drop of 53.38 points, while Llama 4 Maverick demonstrates a significant improvement, gaining 22.30 points.

\begin{table*}[h!]%
\centering%
\small%
\begin{tabular}{@{}l
                S[table-format=1.2]@{\hspace{2pt}}S[table-format=-1.2]
                S[table-format=3.2]@{\hspace{2pt}}S[table-format=-2.2]
                S[table-format=3.2]@{\hspace{2pt}}S[table-format=-2.2]
                S[table-format=2.2]@{\hspace{2pt}}S[table-format=-2.2]
                S[table-format=2.2]@{\hspace{2pt}}S[table-format=-2.2]
                S[table-format=2.2]@{\hspace{2pt}}S[table-format=-2.2]
                @{}}\toprule%
\textbf{Model} & \multicolumn{2}{c}{\textbf{FKGL}} & \multicolumn{2}{c}{\textbf{Sense Aware}} & \multicolumn{2}{c}{\textbf{Multi. Def}} & \multicolumn{2}{c}{\textbf{HeSA}} & \multicolumn{2}{c}{\textbf{Full}} & \multicolumn{2}{c}{\textbf{Both}} \\%
\midrule%
\multicolumn{13}{l}{\textbf{Prompt: Normal}} \\%
Llama 3.1 8B&8.21&\textcolor{own_red}{\tablenum{-0.65}}&\textbf{100.00}&\textcolor{own_green}{\tablenum{+2.70}}&\textbf{100.00}&\textcolor{own_green}{\tablenum{+2.70}}&60.81&\textcolor{own_red}{\tablenum{-7.43}}&52.70&\textit{\textcolor{own_green}{\tablenum{+15.54}}}&29.73&\textcolor{own_green}{\tablenum{+4.05}}\\%
GPT{-}4o mini&8.62&\textcolor{own_red}{\tablenum{-0.36}}&\textbf{100.00}&\textcolor{own_green}{\tablenum{+4.05}}&\textbf{100.00}&\textcolor{own_green}{\tablenum{+4.05}}&10.81&\textcolor{own_red}{\tablenum{-53.38}}&56.08&\textcolor{own_green}{\textit{\tablenum{+6.08}}}&3.38&\textcolor{own_red}{\tablenum{-30.41}}\\%
Qwen3{-}30B A3B&6.51&\textcolor{own_red}{\tablenum{-0.47}}&\textbf{100.00}&\textcolor{own_green}{\tablenum{+4.73}}&99.32&\textit{\textcolor{own_green}{\tablenum{+7.43}}}&\textbf{85.14}&\textcolor{own_green}{\tablenum{+4.73}}&\textbf{58.11}&\textcolor{own_green}{\tablenum{+12.84}}&\textbf{52.03}&\textcolor{own_green}{\tablenum{+12.16}}\\%
Llama 4 Maverick&7.91&\textcolor{own_red}{\tablenum{-1.00}}&\textbf{100.00}&\textcolor{own_green}{\tablenum{+4.05}}&\textbf{100.00}&\textcolor{own_green}{\tablenum{+4.05}}&82.43&\textit{\textcolor{own_green}{\tablenum{+22.30}}}&55.41&\textcolor{own_green}{\tablenum{+8.78}}&41.89&\textit{\textcolor{own_green}{\tablenum{+15.54}}}\\%
DeepSeek v3&7.52&\textcolor{own_red}{\tablenum{-0.87}}&\textbf{100.00}&\textit{\textcolor{own_green}{\tablenum{+6.08}}}&\textbf{100.00}&\textcolor{own_green}{\tablenum{+6.76}}&22.30&\textcolor{own_red}{\tablenum{-1.35}}&66.22&\textcolor{own_green}{\tablenum{+12.16}}&12.16&\textcolor{own_green}{\tablenum{+4.05}}\\%
\midrule%
\multicolumn{13}{l}{\textbf{Prompt: Simple}} \\%
Llama 3.1 8B&6.49&\textcolor{own_red}{\tablenum{-0.34}}&97.30&\textcolor{own_green}{\tablenum{+25.68}}&97.30&\textcolor{own_green}{\tablenum{+26.35}}&15.54&\textcolor{own_red}{\tablenum{-4.05}}&31.08&\textcolor{own_green}{\tablenum{+13.51}}&4.73&\textcolor{own_red}{\tablenum{-0.68}}\\%
GPT{-}4o mini&6.31&\textcolor{own_green}{\tablenum{+0.03}}&95.27&\textcolor{own_green}{\tablenum{+32.43}}&95.27&\textcolor{own_green}{\tablenum{+32.43}}&23.65&\textcolor{own_green}{\tablenum{+16.22}}&38.51&\textcolor{own_green}{\tablenum{+20.95}}&4.73&\textcolor{own_green}{\tablenum{+2.03}}\\%
Qwen3{-}30B A3B&5.18&\textcolor{own_red}{\tablenum{-0.41}}&\textbf{100.00}&\textcolor{own_green}{\tablenum{+15.54}}&\textbf{100.00}&\textcolor{own_green}{\tablenum{+16.22}}&30.41&\textcolor{own_green}{\tablenum{+8.11}}&\textbf{47.97}&\textcolor{own_green}{\tablenum{+16.89}}&12.16&\textcolor{own_green}{\tablenum{+3.38}}\\%
Llama 4 Maverick&6.13&\textcolor{own_red}{\tablenum{-0.69}}&\textbf{100.00}&\textcolor{own_green}{\tablenum{+27.03}}&\textbf{100.00}&\textcolor{own_green}{\tablenum{+28.38}}&\textbf{60.14}&\textit{\textcolor{own_green}{\tablenum{+33.11}}}&\textbf{47.97}&\textit{\textcolor{own_green}{\tablenum{+27.70}}}&\textbf{30.41}&\textit{\textcolor{own_green}{\tablenum{+21.62}}}\\%
DeepSeek v3&5.89&\textcolor{own_red}{\tablenum{-1.15}}&99.32&\textit{\textcolor{own_green}{\tablenum{+37.16}}}&99.32&\textit{\textcolor{own_green}{\tablenum{+37.16}}}&12.84&\textcolor{own_green}{\tablenum{+7.43}}&47.30&\textcolor{own_green}{\tablenum{+27.03}}&4.05&\textcolor{own_green}{\tablenum{+4.05}}\\%
\midrule%
\multicolumn{13}{l}{\textbf{Prompt: ELI5}} \\%
Llama 3.1 8B&2.77&\textcolor{own_red}{\tablenum{-0.35}}&94.59&\textcolor{own_green}{\tablenum{+80.41}}&93.92&\textcolor{own_green}{\tablenum{+82.43}}&25.68&\textcolor{own_green}{\tablenum{+21.62}}&22.30&\textcolor{own_green}{\tablenum{+20.95}}&8.11&\textcolor{own_green}{\tablenum{+7.43}}\\%
GPT{-}4o mini&3.62&\textcolor{own_red}{\tablenum{-0.33}}&85.14&\textcolor{own_green}{\tablenum{+76.35}}&84.46&\textcolor{own_green}{\tablenum{+75.68}}&7.43&\textcolor{own_green}{\tablenum{+7.43}}&24.32&\textcolor{own_green}{\tablenum{+22.97}}&1.35&\textcolor{own_green}{\tablenum{+1.35}}\\%
Qwen3{-}30B A3B&3.21&\textcolor{own_red}{\tablenum{-0.58}}&\textbf{100.00}&\textcolor{own_green}{\tablenum{+62.84}}&\textbf{100.00}&\textcolor{own_green}{\tablenum{+65.54}}&\textbf{35.81}&\textcolor{own_green}{\tablenum{+25.68}}&34.46&\textcolor{own_green}{\tablenum{+25.68}}&10.81&\textcolor{own_green}{\tablenum{+8.11}}\\%
Llama 4 Maverick&2.72&\textcolor{own_red}{\tablenum{-0.26}}&96.62&\textit{\textcolor{own_green}{\tablenum{+86.49}}}&96.62&\textit{\textcolor{own_green}{\tablenum{+87.84}}}&33.11&\textit{\textcolor{own_green}{\tablenum{+30.41}}}&33.78&\textcolor{own_green}{\tablenum{+33.11}}&\textbf{12.84}&\textit{\textcolor{own_green}{\tablenum{+12.84}}}\\%
DeepSeek v3&3.80&\textcolor{own_red}{\tablenum{-0.32}}&97.30&\textcolor{own_green}{\tablenum{+85.81}}&96.62&\textcolor{own_green}{\tablenum{+85.81}}&19.59&\textcolor{own_green}{\tablenum{+18.24}}&\textbf{35.81}&\textit{\textcolor{own_green}{\tablenum{+35.14}}}&5.41&\textcolor{own_green}{\tablenum{+5.41}}\\\bottomrule%
\end{tabular}%
\caption{Performance metrics for models on the HoWN dataset under the Multi-Sense-Aware setting across Normal, Simple, and ELI5 prompt types. Metrics percentage of responses classified as Complete, Full (covering all meanings), HeSA (Helpful Sense Awareness), Both (Full and HeSA), and the average percentage of coarse-grained definitions covered. Deltas are reported to the non-Multi-Sense-Aware setting. coarse-grained definitions covered, with best scores for each prompt type in \textbf{bold}. The top value for each metric, within each prompt type and language, is highlighted in \textbf{bold}. The top delta is highlighted in \textit{italic}.}
\label{tab:multi-sense-aware}
\end{table*}

\section{ML-WIC Analysis}
\label{app:ml-wic}
In Table~\ref{tab:multilang-model-eval}, we present evaluation results on our ML-WIC dataset, including Sense Awareness, the proportion of responses with multiple definitions, and HeSA scores. Our DPO models consistently outperform all other models across every setting.

Table~\ref{tab:fre-score} presents the Flesch Reading Ease (FRE) score \citep{flesch_new_1948}, where higher values indicate simpler language. Scores are reported for English, French, and Russian, as FRE is unavailable for Arabic or Chinese. Across all models, the FRE score increases under simplification constraints, indicating that models adhere to the prompt instructions. We observe an exception with Llama 3.1 8B in the Normal setting for Russian, where the FRE score is negative. This is because the model sometimes produced flawed responses, often repeating words excessively. Additionally, we omit FRE scores for our DPO model, as it occasionally generated responses in English instead of the target language.

Table~\ref{tab:avg-multiple-ml} reports the average number of definitions in responses that contain multiple definitions. A clear trend emerges: Normal yields more definitions on average than Simple, which in turn produces more than ELI5.

\begin{table*}[hbt]
\centering
\small
\setlength{\tabcolsep}{2pt}
\begin{tabular}{
  l
  *{3}{*{5}{r}}
}
\toprule%
\textbf{Prompt / Model} & \multicolumn{5}{c}{\textbf{Sense Aware}} & \multicolumn{5}{c}{\textbf{Multi. Def.}} & \multicolumn{5}{c}{\textbf{HeSA}} \\%
\cmidrule(lr){2%
-%
6}%
\cmidrule(lr){7%
-%
11}%
\cmidrule(lr){12%
-%
16}%
&En&Fr&Ar&Ru&Zh&En&Fr&Ar&Ru&Zh&En&Fr&Ar&Ru&Zh\\%
\midrule%
\multicolumn{16}{l}{\textbf{Prompt: Normal}} \\%
Llama 3.1 8B&\textbf{96.97}&14.97&10.26&6.33&5.10&\textbf{95.96}&12.24&10.26&5.49&4.59&56.23&9.18&0.00&2.95&0.51\\%
DPO Llama 3.1 8B&\textbf{99.66}&\textbf{99.32}&\textbf{99.15}&\textbf{99.16}&\textbf{98.47}&\textbf{99.66}&\textbf{98.30}&\textbf{98.29}&\textbf{97.89}&\textbf{96.94}&\textbf{80.13}&\textbf{95.24}&\textbf{86.32}&\textbf{94.51}&\textbf{53.06}\\%
GPT{-}4o mini&93.94&78.91&92.31&90.30&84.69&93.94&77.89&\textbf{92.31}&89.45&83.67&28.62&17.35&17.09&27.85&16.33\\%
Qwen3{-}30B A3B&94.61&86.39&\textbf{98.29}&89.87&\textbf{100.00}&93.60&84.01&91.45&86.50&\textbf{98.98}&\textbf{65.99}&\textbf{37.41}&\textbf{77.78}&\textbf{42.62}&\textbf{83.16}\\%
Llama 4 Maverick&96.30&54.76&74.36&74.68&45.41&95.29&47.28&73.50&71.73&43.37&47.14&28.91&21.37&31.22&16.84\\%
DeepSeek v3&94.28&\textbf{87.76}&91.45&\textbf{91.56}&87.76&93.60&\textbf{86.39}&91.45&\textbf{90.30}&86.22&20.88&23.13&30.77&18.57&46.94\\%
\midrule%
\multicolumn{16}{l}{\textbf{Prompt: Simple}} \\%
Llama 3.1 8B&64.31&7.48&5.98&2.11&8.67&62.96&5.10&5.98&2.11&8.67&6.06&2.38&0.00&0.00&0.51\\%
DPO Llama 3.1 8B&\textbf{92.93}&\textbf{93.54}&\textbf{96.58}&\textbf{99.58}&\textbf{93.88}&\textbf{92.93}&\textbf{89.12}&\textbf{96.58}&\textbf{98.31}&\textbf{93.37}&\textbf{54.88}&\textbf{69.39}&\textbf{61.54}&\textbf{73.84}&\textbf{31.63}\\%
GPT{-}4o mini&63.30&53.06&76.07&43.88&75.51&63.30&53.06&76.07&43.46&75.51&2.69&0.00&0.00&3.38&1.02\\%
Qwen3{-}30B A3B&\textbf{76.77}&\textbf{59.52}&69.23&\textbf{67.51}&\textbf{82.65}&\textbf{75.76}&\textbf{57.82}&65.81&\textbf{66.24}&\textbf{81.63}&11.45&\textbf{7.48}&\textbf{12.82}&\textbf{11.39}&24.49\\%
Llama 4 Maverick&70.03&28.57&44.44&49.37&68.88&69.36&26.19&44.44&46.84&60.71&\textbf{15.15}&5.78&1.71&8.86&\textbf{26.53}\\%
DeepSeek v3&63.97&47.96&\textbf{80.34}&65.40&73.98&63.97&47.28&\textbf{80.34}&64.14&73.47&2.02&1.70&4.27&8.86&13.27\\%
\midrule%
\multicolumn{16}{l}{\textbf{Prompt: ELI5}} \\%
Llama 3.1 8B&7.07&7.48&0.85&1.27&0.51&6.73&3.40&0.00&0.42&0.51&0.67&4.42&0.85&0.84&0.00\\%
DPO Llama 3.1 8B&\textbf{35.35}&\textbf{35.03}&\textbf{57.26}&\textbf{63.29}&\textbf{34.69}&\textbf{32.66}&\textbf{29.25}&\textbf{51.28}&\textbf{54.43}&\textbf{32.14}&\textbf{17.85}&\textbf{18.71}&\textbf{30.77}&\textbf{42.19}&\textbf{14.80}\\%
GPT{-}4o mini&5.72&6.80&11.11&2.53&6.63&5.72&6.80&10.26&2.11&6.12&0.00&0.34&0.85&0.84&0.51\\%
Qwen3{-}30B A3B&\textbf{22.22}&\textbf{17.35}&11.11&\textbf{14.35}&\textbf{14.29}&\textbf{19.87}&\textbf{11.56}&7.69&\textbf{13.08}&8.16&\textbf{4.71}&\textbf{8.16}&\textbf{4.27}&1.69&\textbf{6.63}\\%
Llama 4 Maverick&10.77&12.93&11.97&9.70&9.69&8.42&10.20&11.97&7.59&6.63&3.70&5.10&0.85&\textbf{2.53}&3.06\\%
DeepSeek v3&8.08&8.16&\textbf{12.82}&8.86&11.73&8.08&7.82&\textbf{12.82}&8.02&\textbf{11.73}&0.00&2.04&0.00&0.84&0.00\\\bottomrule%
\end{tabular}%
\caption{Evaluation scores for Sense Awareness, Multiple Definitions, and HeSA across English (En), French (Fr), Arabic (Ar), Russian (Ru), and Chinese (Zh) for Normal, Simple, and ELI5 prompt types. The top two values for each metric, within each prompt type and language, are highlighted in \textbf{bold}.}
\label{tab:multilang-model-eval}
\end{table*}

\begin{table*}[hbt]%
\centering%
\small%
\setlength{\tabcolsep}{4pt}%
\begin{tabular}{@{}lccccccccc@{}}%
\toprule%
\textbf{Model} & \multicolumn{3}{c}{\textbf{Normal}} & \multicolumn{3}{c}{\textbf{Simple}} & \multicolumn{3}{c}{\textbf{ELI5}} \\%
\cmidrule(lr){2%
-%
4}%
\cmidrule(lr){5%
-%
7}%
\cmidrule(lr){8%
-%
10}%
&En&Fr&Ru&En&Fr&Ru&En&Fr&Ru\\%
\midrule%
Llama 3.1 8B&60.23&41.55&{-}15.62&75.33&75.59&53.07&95.12&92.75&60.16\\%
GPT{-}4o mini&57.72&65.57&37.52&77.20&82.32&58.17&92.91&99.12&79.36\\%
Qwen3{-}30B A3B&60.70&69.99&35.20&78.24&81.34&62.53&92.35&97.60&81.63\\%
Llama 4 Maverick&58.27&61.06&31.13&71.41&65.70&49.44&95.05&94.28&76.11\\%
DeepSeek v3&60.11&68.11&38.64&73.54&77.48&54.26&91.32&97.09&73.84\\\bottomrule%
\end{tabular}%
\caption{Flesch Reading Ease (FRE) scores per prompt type (Normal, Simple, ELI5) across models and languages (English, French, Russian). Higher scores indicate easier-to-read text.}
\label{tab:fre-score}
\end{table*}

\begin{table*}[h!]%
\centering%
\small%
\setlength{\tabcolsep}{4pt}%
\begin{tabular}{@{}lccccccccccccccc@{}}%
\toprule%
\textbf{Model} & \multicolumn{5}{c}{\textbf{Normal}} & \multicolumn{5}{c}{\textbf{Simple}} & \multicolumn{5}{c}{\textbf{ELI5}} \\%
\cmidrule(lr){2%
-%
6}%
\cmidrule(lr){7%
-%
11}%
\cmidrule(lr){12%
-%
16}%
&En&Fr&Ar&Ru&Zh&En&Fr&Ar&Ru&Zh&En&Fr&Ar&Ru&Zh\\%
\midrule%
Llama 3.1 8B&4.80&4.28&2.58&3.85&2.33&2.86&2.13&2.00&2.20&3.35&2.25&2.20&0.00&2.00&3.00\\%
DPO Llama 3.1 8B&5.06&4.95&3.73&5.26&3.67&3.82&3.50&3.19&3.97&3.60&2.81&2.36&2.38&2.97&2.59\\%
GPT{-}4o mini&3.93&3.35&3.11&3.36&3.49&2.32&2.23&2.42&2.23&2.45&2.00&2.10&2.00&2.00&2.08\\%
Qwen3{-}30B A3B&5.60&4.39&3.74&4.66&5.80&2.79&2.65&2.27&3.03&3.77&2.14&2.21&2.11&2.35&2.31\\%
Llama 4 Maverick&4.46&3.63&6.84&3.86&3.40&2.84&2.49&2.81&2.68&3.13&2.12&2.03&2.07&2.22&2.23\\%
DeepSeek v3&4.93&4.23&4.31&4.30&4.56&2.72&2.60&2.87&2.84&3.32&2.00&2.09&2.00&2.05&2.17\\\bottomrule%
\end{tabular}%
\caption{Average number of definitions in responses that contain multiple definitions, reported per prompt type (Normal, Simple, ELI5) across models and languages.}
\label{tab:avg-multiple-ml}
\end{table*}

\begin{figure*}[ht]
\tiny
\centering
\begin{minipage}{\textwidth}
\lstset{backgroundcolor=\color{systembg}} 
\begin{lstlisting}
You are a precise evaluator that determines two flags based on a response explaining a word definition.

Your task:

1. **remark_not_all_listed** -> Set this to True **only if** the response explicitly signals that the list of definitions is incomplete.
   - Set to True when the response includes phrases like:
     - "there are other meanings"
     - "this is just a selection"
     - "some examples include"
     - "among others"
     - "etc."
     - "this is not an exhaustive list"
   - Set to False when:
     - The response lists multiple meanings but gives **no signal** of incompleteness.
     - It introduces uncommon or surprising meanings **but does not suggest there are more**.
     - It uses grouping phrases like "has multiple meanings" or "depends on context" **to summarize the meanings given**, without implying that additional, unlisted meanings exist.
     - It uses playful or contextual comments like "that's a grown-up thing" without signaling incompleteness.

2. **context_clarification_request** -> Set this to True **only if** the response explicitly asks the user to provide more context, clarification, or disambiguation.
   - Set to True when the response includes phrases like:
     - "if you have a specific context in mind, please tell me"
     - "can you specify the context the word is used in?"
   - Set to False when:
     - The response explains multiple meanings or contexts but **does not** request further clarification from the user.
     - It simply states that the word "depends on context" or "has multiple meanings" without asking for more input from the user.
     - It invites the user to interact or reflect (e.g., "Do you have any clips at home?"), but this interaction **is unrelated to clarifying the word's meaning**.

**Example 1**:
Query: What is the definition of "bluff"?
Response: The term "bluff" has multiple definitions depending on the context in which it is used.
1. In geography or geology: ...
2. In poker: ...
3. In general usage: ...
The core idea involves deception or directness, depending on context. Does this make sense?

**Expected Output**:
{'remark_not_all_listed': False, 'context_clarification_request': False, 'explanation': 'remark_not_all_listed is False because the response lists definitions without signaling incompleteness. context_clarification_request is False because no context is requested from the user.'}

**Example 2**:
Query: Explain me "clip" like I am 5 years old.
Response: Ooh, let me explain "clip" to you! ... There are different kinds of clips, like paper clips, hair clips, and even clips for bags or jackets. Isn't that cool? Do you have any clips at home?

**Expected Output**:
{'remark_not_all_listed': False, 'context_clarification_request': False, 'explanation': 'remark_not_all_listed is False because no signal of incompleteness is given. context_clarification_request is False because the interaction is playful, not a request for clarification.'}
\end{lstlisting}
\end{minipage}
\caption{System Prompt used to evaluate Helpful Sense Awareness}
\label{fig:hesa-prompt}
\end{figure*}

\begin{figure*}[ht]
\tiny
\centering
\begin{minipage}{\textwidth}
\lstset{backgroundcolor=\color{userbg}} 
\begin{lstlisting}
You are a helpful assistant. Answer the question as accurately as possible.

**Prompt**: "{query}"

**Model Response**: "{model_response}"
\end{lstlisting}
\end{minipage}
\caption{User prompt layout used in all requests.}
\label{fig:user-prompt}
\end{figure*}

\begin{figure*}[ht]
\tiny
\centering
\begin{minipage}{\textwidth}
\lstset{backgroundcolor=\color{systembg}} 
\begin{lstlisting}
You will extract definitions from the following response and determine whether it expresses one or multiple distinct meanings.

Instructions:

1. Read the response carefully.
2. Identify the distinct meanings or senses of the word being defined.

   - Treat it as **one definition** if the response elaborates, gives examples, describes subtypes, or mentions alternative forms **within the same core meaning**.

   - Split it into **multiple definitions** only if the response clearly and explicitly explains **separate meanings or senses** that differ in their fundamental usage or context. This includes:
     - Different referents (e.g., object vs. concept)
     - Informal, colloquial, or figurative uses that shift the meaning
     - Literal vs. metaphorical senses
     - Different grammatical or functional roles (e.g., noun vs. verb)

   - Do **not** split definitions if:
     - There are multiple examples or elaborations
     - The response describes consequences, comparisons, or related features of the same concept
     - A second meaning is only hinted at or vaguely mentioned without being clearly defined

3. Provide three outputs:

   - 'definitions': a list of distinct definitions (if any).
   - 'category': either 'One' or 'Multiple'.
   - 'explanation': a brief reason for your categorization.

Example 1 (single meaning with details):

Input:
"March means to walk in a formal, organized way as part of a group, often with music and banners."

Output:
definitions:
- To walk in a formal, organized way as part of a group, often with music and banners.
category: One
explanation: This is a single meaning with elaborative details; no distinct senses are presented.

Example 2 (clear multiple meanings):

Input:
"Break can mean to separate into pieces by force, or to pause from an activity."

Output:
definitions:
- To separate into pieces by force.
- To pause from an activity.
category: Multiple
explanation: These are two clearly distinct senses of the word.

Example 3 (no split for subtype and mention):

Input:
"An ass is another kind of animal, also called a donkey. Sometimes people use the word in not-so-nice ways."

Output:
definitions:
- An animal, also called a donkey, similar to a horse but smaller with longer ears.
category: One
explanation: Only one meaning is fully defined; the other usage is mentioned but not explained as a separate sense.

Example 4 (no split for examples and types):

Input:
"A clip is a special tool that helps keep things together, like papers, hair, or bags."

Output:
definitions:
- A special tool that helps keep things together, like papers, hair, or bags.
category: One
explanation: These are examples and subtypes of the same core meaning.
\end{lstlisting}
\end{minipage}
\caption{System prompt used to extract definitions in the Simple and Normal prompting settings}
\label{fig:defintion-prompt}
\end{figure*}

\begin{figure*}[ht]
\tiny
\centering
\begin{minipage}{\textwidth}
\lstset{backgroundcolor=\color{systembg}} 
\begin{lstlisting}
You are a precise evaluator tasked with analyzing a model response defining a word in a child-friendly manner (e.g., prompted with 'Explain me [word] like I am 5 years old'). Your goal is to extract distinct definitions and categorize the response as having one or multiple meanings, recognizing that child-friendly responses often use imaginative or example-heavy language to describe a single core meaning.\n\n
Instructions:\n\n
1. **Read the Response Carefully**:
   - Identify the word being defined and all meanings or senses provided. Note that child-friendly responses often include metaphors, playful language, or multiple examples to engage young audiences.\n\n
2. **Extract Definitions**:
   - **One Definition**: Count as one if the response describes a single core meaning, even if it includes:
     - Imaginative metaphors (e.g., 'a big hug from the water' for a bay).
     - Multiple examples or subtypes (e.g., 'paper clips, hair clips' for a clip).
     - Playful or emotional elaborations (e.g., 'makes you feel like you can fly' for a cape).
     - Descriptions of uses, contexts, or feelings tied to the same concept (e.g., 'worn by superheroes or at Halloween' for a cape).
   - **Multiple Definitions**: Split into separate definitions only if the response **explicitly introduces distinct meanings or senses**, such as:
     - Different referents (e.g., a physical object vs. an abstract concept).
     - Literal vs. metaphorical senses with clear shifts in meaning.
     - Different grammatical roles (e.g., noun vs. verb) with distinct implications.
   - **Do Not Split**:
     - Examples, subtypes, or contexts that support the same core concept (e.g., 'playing soccer or with toys' for play).
     - Stylistic, emotional, or imaginative variations that do not change the core sense (e.g., 'a magic cloak' for a cape).
   - **Compound Sentences or 'or'-Phrases**: Evaluate carefully. Only split if they describe distinct senses with different implications (e.g., 'to put or adjust' may be two senses if contexts differ significantly).
   - **Kid-Friendly Language**: Treat metaphors, analogies, or engaging descriptions as elaborations of a single meaning unless they clearly introduce a new sense. For example, 'a big hug from the water' and 'calm place for boats' describe the same geographic feature (bay).
   - Do not verify factual correctness-extract only what is present.\n\n
3. **Assign Category**:
   - **None**: No definitions provided.
   - **One**: One distinct definition extracted.
   - **Multiple**: Two or more distinct definitions extracted.\n\n
4. **Provide Outputs**:
   - `definitions`: A list of distinct definitions, summarizing each sense concisely.
   - `category`: 'None', 'One', or 'Multiple'.
   - `explanation`: A brief reason for the categorization, addressing why details are treated as elaborations or distinct senses in the child-friendly context.\n\n
Example 1 (Single Meaning with Kid-Friendly Elaborations):
Input: 'A bay is a special area where the ocean curves into the land. It's like a big hug from the water! Bays are good places for boats because the water is calmer and protected from big ocean waves.'
Output:
{
  'definitions': ['A special area where the ocean curves into the land, providing a calm place protected from big waves'],
  'category': 'One',
  'explanation': 'The response describes one meaning of bay (a geographic feature) with kid-friendly metaphors ('big hug') and elaborations (calm water, boats). No distinct senses are introduced.'
}\n\n
Example 2 (Single Meaning with Playful Language):
Input: 'A cape is a special kind of clothing that you wear on your back! It's like a big piece of fabric that goes around your neck and hangs down behind you. Imagine you're a superhero, like Superman or Batman! You wear a cape to make you feel strong and brave. It's like a magic cloak that helps you fly (well, not really, but it makes you feel like you can!). Some people wear capes on Halloween or at costume parties to dress up like their favorite superheroes.'
Output:
{
  'definitions': ['A special kind of clothing that you wear on your back, like a big piece of fabric that goes around your neck and hangs down behind you'],
  'category': 'One',
  'explanation': 'The response defines cape as clothing with kid-friendly elaborations (superhero imagery, Halloween use, emotional effects). These are not distinct senses but support the same concept.'
}\n\n
Example 3 (Single Meaning with Examples):
Input: 'Ooh, let me explain "clip" to you! A clip is a special tool that helps us keep things together. It's like a little helper! You can use a clip to hold papers or keep hair in place. There are different kinds of clips, like paper clips or hair clips.'
Output:
{
  'definitions': ['A special tool that helps keep things together, like for papers or hair'],
  'category': 'One',
  'explanation': 'The response defines clip as a fastening tool with kid-friendly examples (papers, hair). These are elaborations of one meaning, not distinct senses.'
}\n\n
Example 4 (Multiple Meanings, Hypothetical):
Input: 'A bank is a place where you keep your money, like where your piggy bank savings go! It's also a piece of land next to a river, like where you might have a picnic by the water.'
Output:
{
  'definitions': ['A place where money is kept, like a financial institution', 'A piece of land next to a river'],
  'category': 'Multiple',
  'explanation': 'The response provides two distinct meanings of bank (financial institution and river land), each with kid-friendly descriptions. These are separate senses with different referents.'
}\n\n
Example 5 (Single Meaning with Cultural Context):
Input: 'Oh boy, are you going to love learning about the Cheyenne! So, a long, long time ago, there was a group of Native American people called the Cheyenne. They lived in a place called the Great Plains, which is a big area of grasslands in the middle of the United States. The Cheyenne people were very good at riding horses and hunting animals like buffalo.'
Output:
{
  'definitions': ['A Native American tribe that lived in the Great Plains'],
  'category': 'One',
  'explanation': 'The response defines Cheyenne as a Native American tribe with kid-friendly details (Great Plains, horses, hunting). These are elaborations of one meaning, not distinct senses.'
}
\end{lstlisting}
\end{minipage}
\caption{System prompt used to extract definitions in the ELI5 prompting setting}
\label{fig:eli5-prompt}
\end{figure*}

\begin{table*}[t]
    \includegraphics[width=\linewidth]{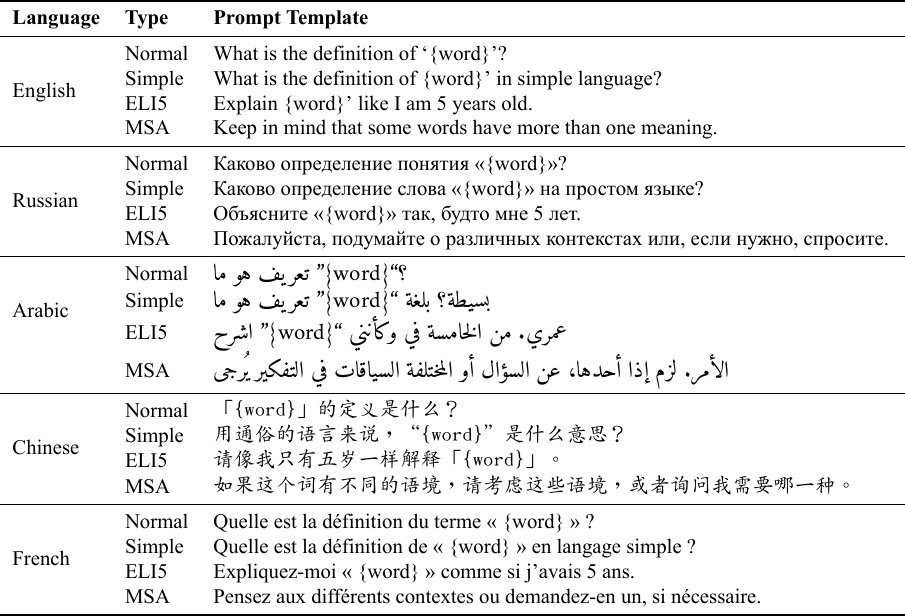}
    \caption{Used prompt types across all languages. Multi-Sense-Aware (MSA) follows any of the other prompts.}
    \label{fig:language-prompts}
\end{table*}

\end{document}